\documentclass[preprint,12pt,authoryear]{elsarticle}

\usepackage{amssymb}
\usepackage{graphicx}
\usepackage{multirow}
\usepackage{booktabs}
\usepackage{array} 
\usepackage[margin=1in]{geometry}

\usepackage[normalem]{ulem}

\usepackage{amsmath}
\usepackage[hidelinks]{hyperref}
\newcolumntype{C}[1]{>{\centering\arraybackslash}p{#1}}

\begin{document}

\begin{frontmatter}

\title{SpecAware: A Spectral-Content Aware Foundation Model for Unifying Multi-Sensor Learning in Hyperspectral Remote Sensing Mapping}

\author[a,c,d]{Renjie Ji\fnref{fn1}}
\author[a,c,d]{Xue Wang\fnref{fn1}}
\fntext[fn1]{Renjie Ji and Xue Wang contributed equally to this work.}
\author[a,c,d]{Chao Niu}
\author[e,a,b]{Wen Zhang}
\author[f]{Yong Mei}
\author[a,b,c]{Kun Tan\corref{mycorrespondingauthor}}
\cortext[mycorrespondingauthor]{Corresponding author}
\ead{tankuncu@gmail.com}

\affiliation[a]{
            organization={Key Laboratory of Spatial-Temporal Big Data Analysis and Application of Natural Resources in Megacities (Ministry of Natural Resources), East China Normal University},
            city={Shanghai},
            postcode={200241},
            country={China}}
\affiliation[b]{
            organization={School of Geospatial Artificial Intelligence, East China Normal University},
            city={Shanghai},
            postcode={200241},
            country={China}}            
\affiliation[c]{
            organization={Key Laboratory of Geographic Information Science (Ministry of Education), East China Normal University},
            city={Shanghai},
            postcode={200241},
            country={China}}
\affiliation[d]{
            organization={School of Geographic Sciences, East China Normal University},
            city={Shanghai},
            postcode={200241},
            country={China}}
\affiliation[e]{
            organization={Shanghai Municipal Institute of Surveying and Mapping},
            city={Shanghai},
            postcode={200063},
            country={China}}
\affiliation[f]{
            organization={Institute of Defense Engineering, AMS},
            city={Beijing},
            postcode={100036},
            country={China}}

\begin{abstract}
Hyperspectral imaging (HSI) is a critical technique for fine-grained land-use and land-cover (LULC) mapping. However, the inherent heterogeneity of HSI data, particularly the variation in spectral channels across sensors, has long constrained the development of model generalization via transfer learning or joint training. Existing HSI foundation models show promise for different downstream tasks, but typically underutilize the critical guiding role of sensor meta-attributes and image semantic features, resulting in limited adaptability to cross-sensor joint learning. To address these issues, we propose SpecAware, which is a novel hyperspectral spectral-content aware foundation model for unifying multi-sensor learning for HSI mapping. To support this work, we constructed the Hyper-400K dataset, which is a new large-scale pre-training dataset with over 400\,k high-quality patches from diverse airborne AVIRIS sensors that cover two data processing levels (L1 and L2). The core of SpecAware is a hypernetwork-driven unified image embedding process for HSI data. Firstly, we designed a meta-content aware module to generate a unique conditional input for each HSI sample, tailored to each spectral band by fusing the sensor meta-attributes and its own image content. Secondly, we designed the HyperEmbedding module, where a sample-conditioned hypernetwork dynamically generates a pair of matrix factors for channel-wise encoding. This process implements two-step matrix factorization, consisting of adaptive spatial pattern extraction and latent semantic feature projection, yielding a unified hyperspectral token representation. Thus, SpecAware learns to capture and interpret spatial-spectral features across diverse scenes and sensors, enabling adaptive processing of variable spectral channels within a unified multi-sensor joint pre-training framework. Extensive experiments on seven datasets demonstrate that SpecAware can learn competitive feature representations, compared to the existing pre-trained models, achieving superior performance in a suite of LULC tasks, including large-scale land-cover semantic segmentation, change detection, and scene classification. The proposed SpecAware model will be released at \href{https://github.com/busbyjrj/SpecAware}{https://github.com/busbyjrj/SpecAware}.
\end{abstract}


\end{frontmatter}

\section{Introduction}

\begin{figure}[tbp]
\vspace{-1.0em}
\centering
\includegraphics[width=0.95\linewidth]{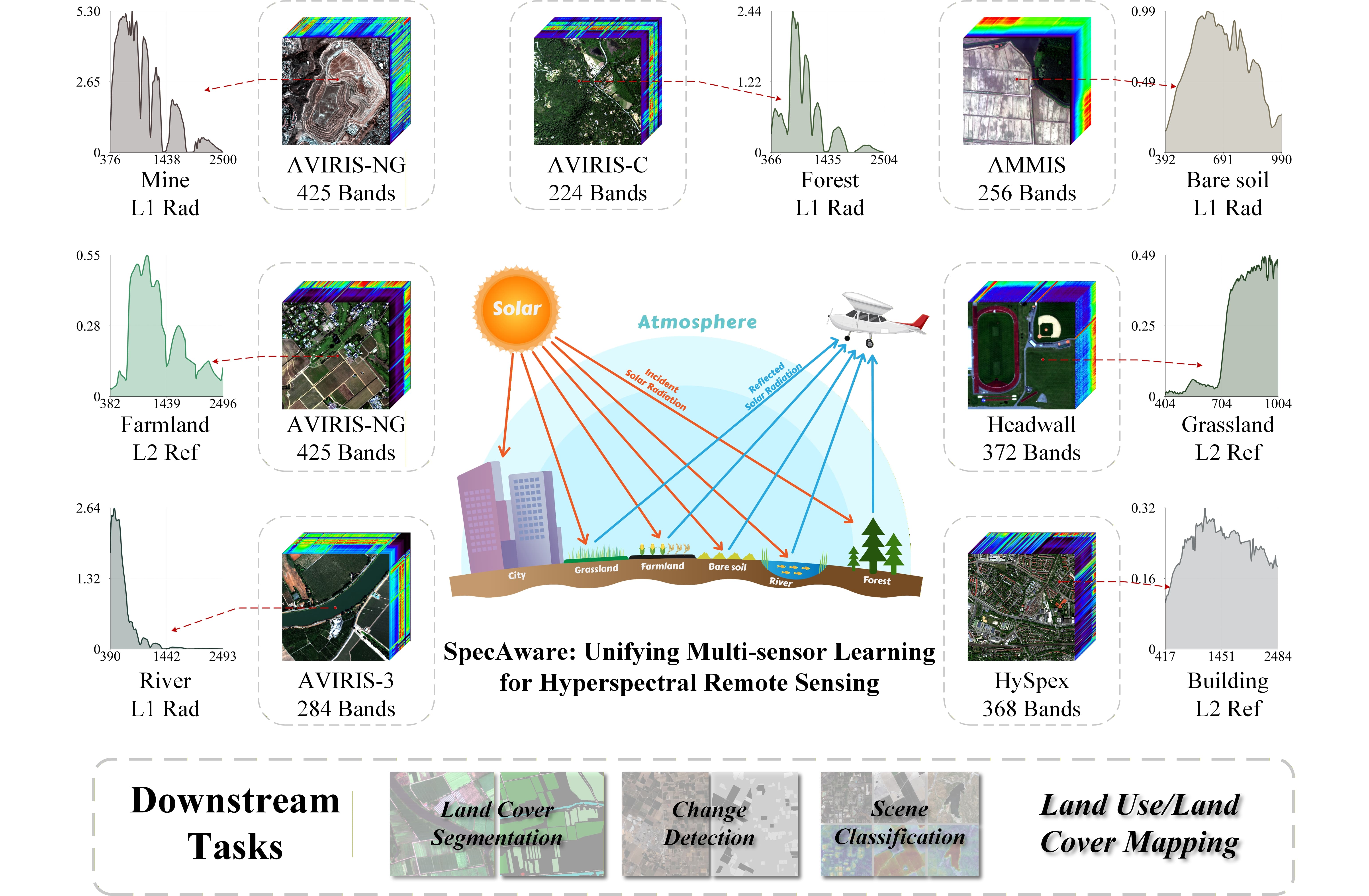}
\caption{The diversity of hyperspectral sensors and data scenes.}
\label{fig:GA}
\vspace{-1.0em}
\end{figure}

Hyperspectral imaging (HSI) has emerged as a powerful and transformative technology for observing the Earth's surface, especially for fine-grained land-use and land-cover (LULC) mapping \citep{wambuguHyperspectralImageClassification2021, thoreauToulouseHyperspectralData2024}. By acquiring data across hundreds of contiguous, narrow spectral bands, HSI enables the construction of a detailed spectral signature for each pixel \citep{paolettiDeepLearningClassifiers2019}, with airborne platforms being particularly advantageous by synergistically combining this detailed spectral information with a very high spatial resolution (Figure~\ref{fig:GA}). This rich, dual-faceted spatial-spectral information has made HSI data invaluable for precision LULC applications, such as crop cultivation monitoring \citep{zhongWHUHiUAVborneHyperspectral2020}, fine-grained land-cover classification \citep{jiPatchOutNovelPatchfree2025}, and tree species classification \citep{jiaInfluenceBRDFEffects2024}.

As shown in Figure~\ref{fig:GA}, airborne HSI data, characterized by the high dimensionality, exhibit many variations, including an inconsistent spatial resolution, differing spectral ranges and resolutions, and different data processing levels (e.g., Level-1 (L1) radiance, Level-2 (L2) reflectance) to suit different application tasks. These variations require interpretation models for LULC tasks to possess both high computational performance and robust feature extraction capabilities. Recently, deep learning approaches have achieved significant advantages in HSI analysis. Architectures such as convolutional neural networks (CNNs) \citep{wangUnifiedMultiscaleLearning2022}, Transformers \citep{jiPatchOutNovelPatchfree2025}, and Mamba models \citep{liMambaHSISpatialSpectral2024} have been successfully applied to HSI LULC tasks, including classification \citep{kumarDeepLearningHyperspectral2024}, change detection \citep{wangSpectralSpatialTemporal2022} and scene classification \citep{liSSFNetSpectralSpatial2025}, with notable success.

However, most of the aforementioned models are trained in a fully supervised manner, which necessitates a large number of labeled training samples. While the acquisition of HSI data has become relatively straightforward with the advancement of HSI sensor technology, the process of annotating these samples with accurate LULC labels remain a difficult and labor-intensive task \citep{jiPatchOutNovelPatchfree2025}. Furthermore, due to the differences in ground objects and HSI sensors, it is difficult to directly apply the valuable HSI labels to new scenes \citep{huangCrosssceneWetlandMapping2023, wangUnifiedMultiscaleLearning2022}. Several studies have highlighted that the limited availability of labeled HSI samples constrains the further development of advanced models \citep{schmittThereAreNo2023}. To overcome the limitation of scarce labeled HSI samples, researchers have developed various techniques to enhance model performance, e.g., data augmentation \citep{zhengFPGAFastPatchFree2020}, transfer learning \citep{wangUnifiedMultiscaleLearning2022}, semi-supervised learning \citep{yaoPseudolabelBasedUnreliableSample2023}, and weakly supervised learning \citep{yangITERImagetoPixelRepresentation2024}. However, the potential learnable features within the vast quantities of HSI data lacking specific land-use labels remain largely untapped.

In recent years, self-supervised learning (SSL) offers a powerful pre-training paradigm without labels, primarily through contrastive learning \citep{houHyperspectralImageryClassification2022, ouHyperspectralImageChange2022} and masked image modeling (MIM) \citep{heMaskedAutoencodersAre2022}. In the field of remote sensing, the dual challenges of the high data annotation costs and the vast abundance of unlabeled data have spurred significant interest in SSL. The contrastive methods typically learn global, image-level features, while the MIM approach reconstructs fine-grained details from masked inputs, making it better suited for pixel-dense remote sensing tasks such as land-cover classification and change detection \citep{tu2HM22024}.

In detail, the MIM-based methods, exemplified by the popular masked autoencoder (MAE) \citep{heMaskedAutoencodersAre2022}, operate by partitioning an image into patches, masking a random subset, and training a model to predict the content of the masked regions from the visible ones. Over the past three years, MAE-based models have rapidly advanced in remote sensing, particularly in multispectral remote sensing, with models such as SpectralGPT \citep{hongSpectralGPTSpectralRemote2024}, SatMAE \citep{congSatmaePretrainingTransformers2022}, and SatMAE++ \citep{nomanRethinkingTransformersPretraining2024}. For HSI, research based on MAE-style SSL has emerged. Specifically, HyperSIGMA is the first large-scale vision Transformer (ViT)-based foundation model designed to create a unified method for interpreting HSI data across various tasks and scenes \citep{wangHypersigmaHyperspectralIntelligence2025}. To avoid the limitations imposed by data from a single specific scenario, the large-scale HyperGlobal-450K dataset was designed \citep{wangHypersigmaHyperspectralIntelligence2025}. Similarly, SpectralEarth, which is a large spectral foundation model, was pre-trained on 3.3 TB (540k images) of EnMAP satellite data, and introduces a spectral adapter module to facilitate adaptation to other HSI data \citep{brahamSpectralEarthTrainingHyperspectral2025}. 
FactoFormer utilizes novel factorized spectral and spatial Transformers to address the underutilization of spatial information and the data scarcity in HSI Transformers \citep{mohamedFactoFormerFactorizedHyperspectral2024}. S\textsuperscript{2}HM\textsuperscript{2} incorporates a 3D masking strategy and a spectral-spatial consistency loss, enabling it to extract spectral and spatial characteristics from HSI data \citep{tu2HM22024}. HyperSL employs a point-wise pre-training strategy that standardizes spectral vectors into unified tokens and integrates wavelength-aware positional embeddings, thereby enabling uniform processing and feature alignment for data from various sensors \citep{kongHyperSLSpectralFoundation2025}. While MAE-based SSL has shown promise for HSI data, significant challenges remain in adapting it to the high dimensionality, diverse sensors, and varying imaging conditions inherent in HSI data.

Firstly, developing generalizable LULC models that can operate across different scenes is severely hampered by the imagery's inherent complexity and diversity. This heterogeneity arises from the significant variations in both the inter-sensor parameters (e.g., spectral range and resolution) and intra-sensor conditions (e.g., atmospheric effects, channel failures). Therefore, there is a need for pre-trained models with the flexibility to handle varied sensor characteristics and dynamic spectral band configurations.

Secondly, tokenizing high-dimensional HSI data for ViT-based pre-training presents a unique challenge. The naive approach of using a simple linear layer to project high-dimensional data risks a significant loss of critical spectral information. However, more powerful methods, such as separate pre-training for spectral and spatial features, can increase the computational burden. Therefore, this necessitates a tokenization strategy for HSI data that is both computationally efficient and spectrally aware.

Thirdly, the scale and diversity of training data present another major challenge for HSI pre-training. Early models were constrained by small datasets that ultimately impair the generalization and computational efficiency \citep{mohamedFactoFormerFactorizedHyperspectral2024, wangHSIMAEUnifiedMasked2024}. Recent large-scale datasets, e.g., SpectralEarth \citep{brahamSpectralEarthTrainingHyperspectral2025} and HyperGlobal-450K \citep{wangHypersigmaHyperspectralIntelligence2025}, have greatly expanded the data availability. However, the satellite-based datasets lack a high spatial resolution.
Hyper-Seg \citep{liHyperFreeChanneladaptiveTuningfree} leverages high-resolution airborne HSI data from AVIRIS-C sensor, yet its coverage is restricted to L1 radiance products only.
Currently, the proliferation of different hyperspectral sensors presents a clear opportunity to establish a large-scale, multi-sensor aerial HSI pre-training benchmark.

To address the aforementioned challenges, we propose the SpecAware framework, which is a spectral-content aware foundation model for unifying multi-sensor learning in hyperspectral imagery mapping. The core objective of SpecAware is to achieve adaptive HSI data interpretation across diverse sensors and scenes. The framework consists of three main stages. Firstly, it performs a joint encoding of the HSI meta-attributes and content features. A novel meta-content aware dual-driven encoder is designed to learn how factors such as land-cover type and sensor properties influence the spectral and spatial characteristics in HSI data, thereby aligning the attribute-aware and content-aware representations. 
The second stage focuses on spectral-spatial property image embedding. We introduce a flexible hypernetwork, driven by the encoded representations from the first stage, to generate two matrix factors for each spectral channel of the HSI data, enabling a unified, channel-wise spatial-spectral embedding process. The third stage then utilizes a similar hypernetwork-based reconstruction layer to restore the masked spatial-spectral information. The main contributions of this work are as follows:

1) We developed an flexible hypernetwork, HyperEmbedding, to generate matrix weights. Through a two-step matrix operation involving adaptive spatial pattern extraction and latent semantic feature projection, the proposed SpecAware framework can effectively process variable spectral channels and establish a unified multi-sensor learning framework for diverse HSI sources. In downstream tasks, the HyperEmbedding module requires no architectural changes to accommodate spectral configurations of unseen hyperspectral sensors.

2) We proposed a meta information and feature content aware dual-driven encoder. This encoder integrates the spectral metadata and image content to generate highly contextualized sample features. By generating a dedicated conditional token for each spectral channel, the encoder fuses heterogeneous information into a unified spectral fingerprint, which provides informative conditioning signals to the hypernetwork and thereby enables adaptive embedding of diverse multi-source HSI data.

3) We constructed a large-scale, high-resolution airborne HSI pre-training dataset, Hyper-400K, covering three generations of AVIRIS sensors and two processing levels.
Based on this dataset, we pretrained the SpecAware hyperspectral foundation model to obtain high-quality spatial-spectral representations. In the downstream LULC tasks, SpecAware demonstrates robust generalization, including land-cover semantic segmentation, change detection, and scene classification.

\section{Related Work}

\subsection{Supervised deep learning methods for HSI interpretation}
Most CNN- or Transformer-based models such as A\textsuperscript{2}S\textsuperscript{2}K-ResNet \citep{royAttentionBasedAdaptiveSpectral2021}, SpectralFormer \citep{hongSpectralFormerRethinkingHyperspectral2022}, or PASSNet \citep{jiPASSNetSpatialSpectral2023} leverage 2D or 3D convolutional layers and multi-head self-attention to exploit the rich spatial and spectral features within hyperspectral imagery. However, these models operate in a patch-wise manner, processing individual image patches to predict the label of the central pixel. While this approach is efficient and feasible for small-scale scenes, it suffers from limited computational efficiency and significant redundant calculations when applied to large-scale LULC image interpretation tasks \citep{jiPatchOutNovelPatchfree2025}. The introduction of the fast patch-free global learning (FPGA) framework, which is a fully end-to-end classification network for HSI data, has enabled hyperspectral semantic segmentation tasks \citep{zhengFPGAFastPatchFree2020}. Follow-up studies have also validated this patch-free approach for large-scale imagery, in methods such as UML \citep{wangUnifiedMultiscaleLearning2022}, PatchOut \citep{jiPatchOutNovelPatchfree2025}, UM2Former \citep{xuUM2FormerUShapedMultimixed2025}, GlobalMind \citep{huGlobalMindGlobalMultihead2024}, etc.

\subsection{HSI data pre-trained and foundation models}

The introduction of MAE models has greatly advanced the development of vision foundation models. However, for hyperspectral remote sensing, a new set of challenges has also been presented. To handle the rich spectral and spatial features of HSI data, some methods, such as SS-MAE \citep{linSSMAESpatialSpectral2023} and HyperSIGMA \citep{wangHypersigmaHyperspectralIntelligence2025}, independently pre-train on spatial and spectral features, with SS-MAE adding a lightweight CNN and HyperSIGMA using sparse attention to reduce computation cost. However, this decoupled pre-training strategy leads to prolonged pre-training times and complicates the model usage. S\textsuperscript{2}HM\textsuperscript{2} adopts a 3D masking strategy, aiming to jointly learn the spectral and spatial features in a single pre-training stage \citep{tu2HM22024}. However, differing from video data, due to the highly consistent spatial structure across the spectral bands of HSI data, the model may easily reconstruct the spatial information.

Furthermore, achieving compatibility with HSI data from diverse sensors, without requiring modifications to the model's architecture, has long been a significant challenge. By pre-training on randomly selected 100-band contiguous subsets, HyperSIGMA focuses on learning the internal relationships within the data, thereby rendering the model agnostic to absolute wavelength positions \citep{wangHypersigmaHyperspectralIntelligence2025}. SpectralEarth incorporates a 3D convolution-based spectral adapter module to map the input HSI data to a fixed number of channels; however, it remains agnostic to the physical meaning of the wavelengths \citep{brahamSpectralEarthTrainingHyperspectral2025}. HyperFree, which was presented as the first foundation model based on prompt engineering, establishes a library of band weights where the image encoding weights are selected by the matching wavelength values, but its effectiveness is dependent on setting appropriate intervals based on the sensor’s spectral resolution, to avoid leaving some wavelengths unmatched \citep{liHyperFreeChanneladaptiveTuningfree}. 

\subsection{Hypernetworks}

A fundamental limitation of deep learning models is that, once trained, their architecture and parameter shape are fixed \citep{chauhanBriefReviewHypernetworks2024}, which creates a significant hurdle when adapting to diverse datasets such as HSI datasets with their variable channel structures. Hypernetworks address this rigidity by functioning as a meta-model that generates the weights for a separate target network \citep{chauhanBriefReviewHypernetworks2024, haHyperNetworks2017}. This allows for dynamic adaptation, as both models are optimized jointly through an end-to-end, differentiable training procedure.

\begin{figure}[htbp]
    \centering
    \includegraphics[width=0.7\linewidth]{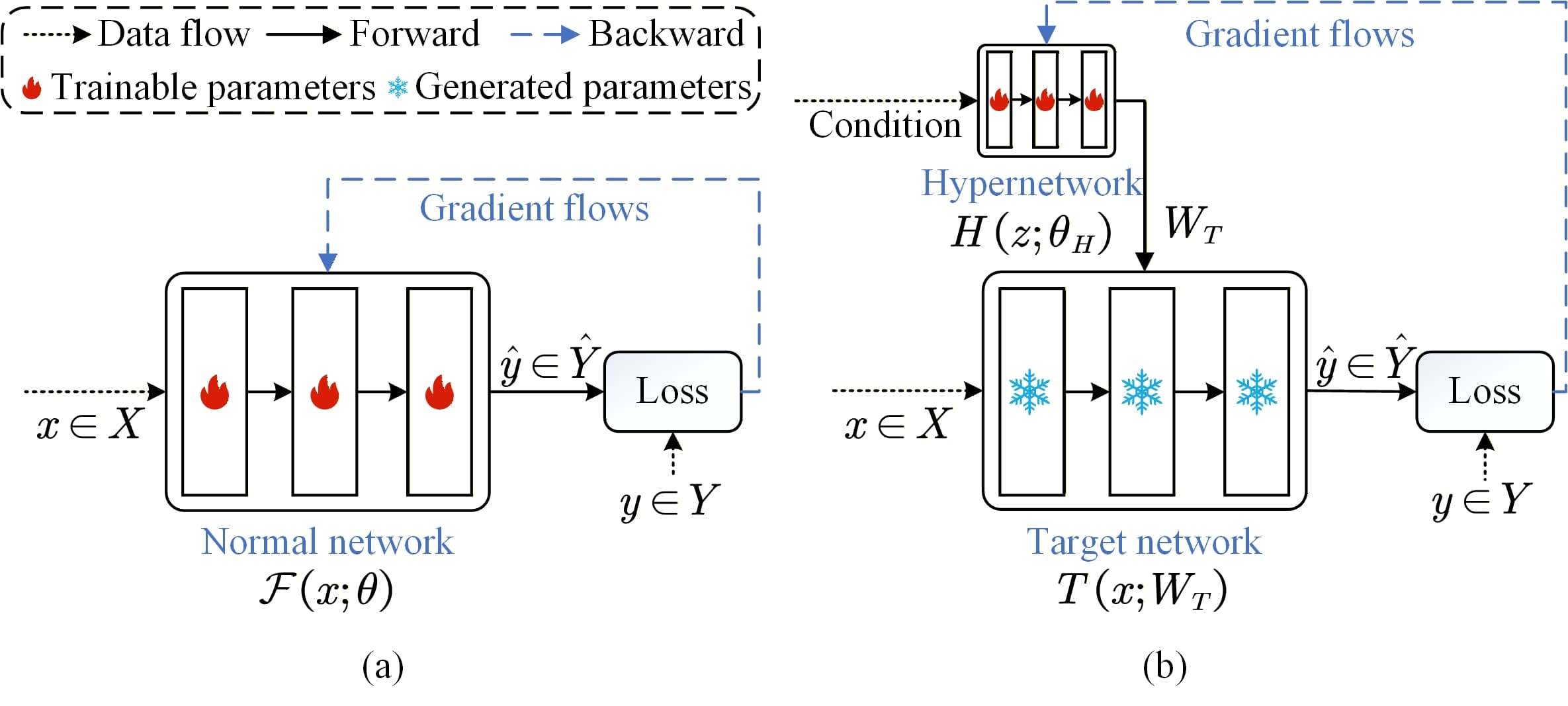}
    \caption{Architectural comparison between (a) a conventional neural network and (b) a hypernetwork.}
    \label{fig:sub-ab}
\end{figure}

A hypernetwork is a neural network whose main function is to generate the weights or parameters for a separate, main neural network \citep{chauhanBriefReviewHypernetworks2024, haHyperNetworks2017}. Formally, as shown in Figure~\ref{fig:sub-ab}, it consists of two key components: the Target network (denoted as $T$), which is the main network designed to solve a specific task, and the Hypernetwork (denoted as $H$), whose task is to generate the weights for the target network. The generation process is conditioned on an input vector $z$, which encapsulates the context for the desired task or behavior. The core relationship can be expressed as:
\begin{equation}
W_T=H\left(z;\theta_H\right)
\end{equation}
\begin{equation}
y=T\left(x;W_T\right)=T\left(x;H\left(z;\theta_H\right)\right)
\end{equation}where $\theta_H$ represents the parameters of the hypernetwork $H$ itself, which are learned during the end-to-end training process, and $W_T$ represents the weights for the target network $T$.

Thus, the hypernetwork $H$ acts as a dynamic weight generator, producing a tailored target network $T$ for the given conditions specified by the input $z$, allowing it to adapt to variable input channels and data properties. Furthermore, a hypernetwork can achieve weight compression by requiring fewer trainable parameters than a corresponding fixed-weight module \citep{karimimahabadiParameterefficientMultitaskFinetuning2021}, making it particularly effective for high-dimensional data. 

Hypernetworks have achieved impressive results in numerous areas, including generative models \citep{dinhHyperInverterImprovingStyleGAN2022} and style transfer \citep{sunFlexibleStyleTransfer2025}. The application of hypernetworks has also extended to the field of remote sensing interpretation. For instance, Hyper-SR dynamically generates the weights for an existing super-resolution network, accelerating the model convergence with only a marginal 0.12\% increase in computational cost \citep{mishraHyperSRHypernetworkBased2024}. To address the challenge of scarce remote sensing data for lakes, a multi-mask, metadata-driven hypernetwork approach was proposed, which effectively leverages the available metadata to improve the model accuracy \citep{graffeuilleRemoteSensingWater2024}. Recently, to process large-scale, multimodal Earth observation data, the DOFA model employs a hypernetwork that leverages the wavelength as a unified parameter to adapt to various spectral modalities \citep{xiongNeuralPlasticityInspiredMultimodal2024}. 

Considering the complex diversity in HSI data that arises from the significant differences between sensors, data processing levels, atmospheric conditions, seasons, and geographical locations, to address the problem and enable cross-sensor generalization, we propose a novel pre-training framework for large-scale multi-sensor data, which features a hypernetwork jointly driven by both metadata and image features.

\section{Datasets}

\subsection{Data sources}

The combination of a high spectral and high spatial resolution gives airborne HSI data a broad range of applications \citep{jiPatchOutNovelPatchfree2025, zhongWHUHiUAVborneHyperspectral2020}. However, the majority of HSI pretraining datasets are derived from satellite-borne sensors, as can be seen in Table~\ref{tab:pretraining_data}, which presents a domain gap, as the coarse spatial resolution of satellite data is insufficient to meet the demands of high-resolution airborne applications. 
In recognition of this, we selected open data from the prominent AVIRIS sensors for this research. Data acquired by the AVIRIS sensors have been used to create many important HSI datasets \citep{paolettiDeepLearningClassifiers2019}, such as the Indian Pines, Salinas, and San Diego datasets. In this study, in order to further investigate the capabilities of HSI sensors, we utilized data from the three generations of AVIRIS sensors—AVIRIS Classic \citep{hamlinImagingSpectrometerScience2011}, AVIRIS-NG \citep{chapmanSpectralRadiometricCalibration2019}, and AVIRIS-3 \citep{eckertAVIRIS3NextGenerationImaging2024}—all of which are publicly available from NASA\footnotemark[1]\footnotemark[2]. 
These three generations of sensors provide 224, 425, and 284 spectral bands, respectively, spanning the 400\,\text{--}\,2500 nm spectral range from the visible to the shortwave infrared.

\footnotetext[1]{\href{https://aviris.jpl.nasa.gov/}{https://aviris.jpl.nasa.gov/}}
\footnotetext[2]{\href{https://avirisng.jpl.nasa.gov/}{https://avirisng.jpl.nasa.gov/}}

\begin{table}[htbp]
\centering
\caption{Summary of the pre-training data for the HSI foundation models.}
\label{tab:pretraining_data}
\scriptsize
\begin{tabular}{lcccccc} 
\toprule
Dataset & \# Images & Patch size & GSD (m) & \# Bands & Data level & Sensor \\ 
\hline
HySpecNet-11k \\ \citep{fuchsHySpecNet11kLargeScaleHyperspectral2023} & $\sim$11\,k & 128 $\times$ 128 & 30 & 202 & L2 & EnMAP \\
HSIHybrid \\ \citep{wangHSIMAEUnifiedMasked2024} & $\sim$4\,m & 9 $\times$ 9 & — & — & — & — \\
HyperGlobal-450K \\ \citep{wangHypersigmaHyperspectralIntelligence2025} & $\sim$447\,k & 64 $\times$ 64 & 30 & 175/150 & L1 & EO-1/GF-5B \\
SpectralEarth \\ \citep{brahamSpectralEarthTrainingHyperspectral2025} & $\sim$538\,k & 128 $\times$ 128 & 30 & 202 & L2 & EnMAP \\
HyperSL \\ \citep{kongHyperSLSpectralFoundation2025} & $\sim$300\,m & 1 $\times$ 1 & 30/30 & 217/229 & L2 & EnMAP/DESIS \\
Hyper-Seg \\ \citep{liHyperFreeChanneladaptiveTuningfree} & $\sim$42\,k & $512\,\times\,512$ & $0.6\,\text{--}\,5.0$ & 224 & L1 & AVIRIS-C \\
\bottomrule
\end{tabular}
\end{table}

\subsection{Data acquisition and processing}

\begin{figure}[htbp]
    \centering
    \includegraphics[width=0.8\linewidth]{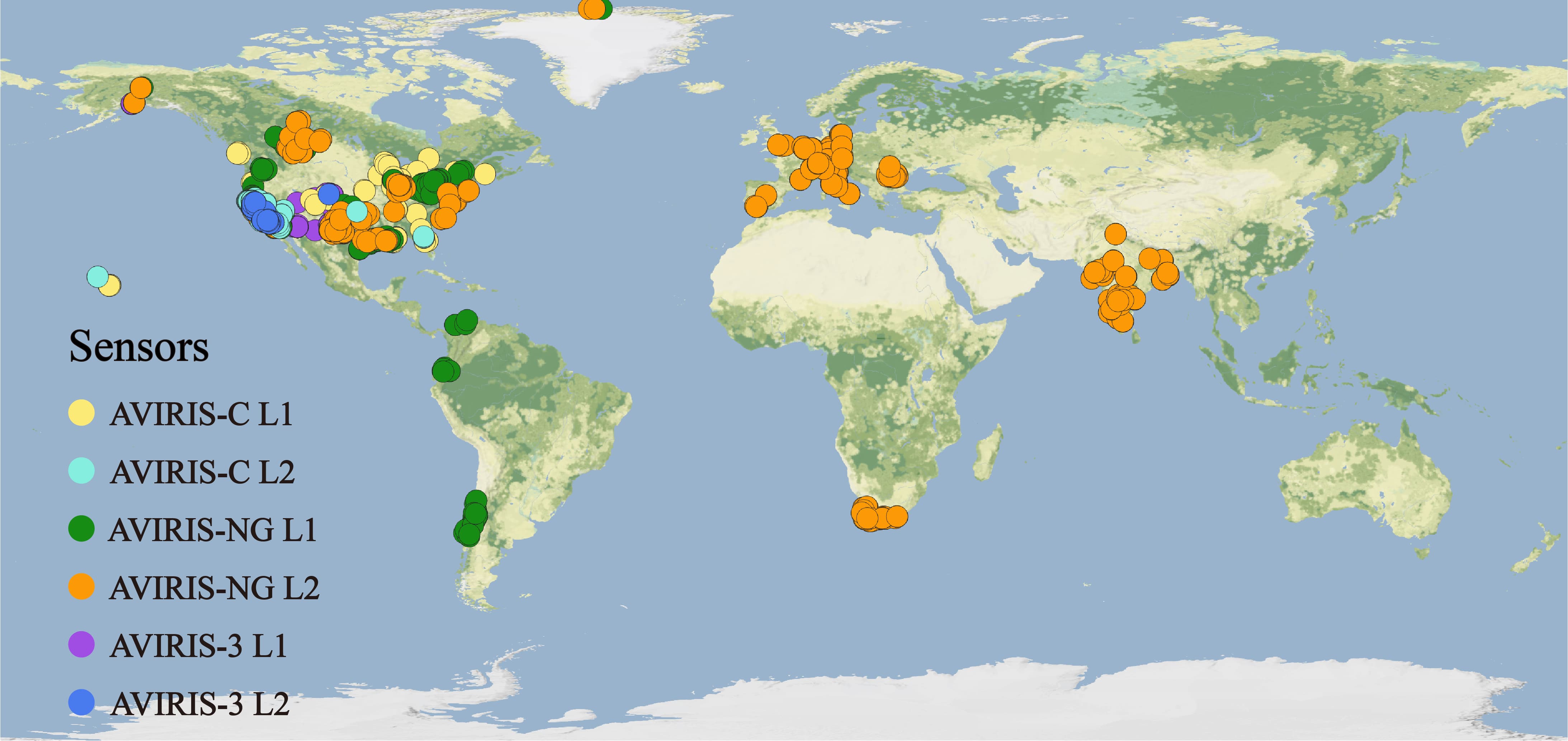}
    \caption{
    Geographic distribution of the Hyper-400K dataset. Most AVIRIS acquisitions are concentrated in North America, particularly the western United States, according to their flight campaigns. Additional coverage comes from the west coast of South America, southern Africa, Europe, and India.
    }
    \label{fig:hyper_400k}
\end{figure}

The Hyper-400K dataset was constructed from AVIRIS airborne HSI collected over $2006\,\text{--}\,2025$, covering diverse surface types, including urban areas, farmland, grassland, forests, mountains, lakes and bare soil.
Figure \ref{fig:hyper_400k} illustrates the geographic distribution of the Hyper-400K dataset. To address the limited diversity in existing pre-training datasets, both L1 calibrated radiance and L2 surface reflectance products from three AVIRIS sensor generations were included in this study. This dataset has undergone rigorous quality filtering to facilitate the learning of spatial and spectral features with high practical application value.

Firstly, based on the AVIRIS official flight plans and quick view, we performed rigorous screening to remove images with low quality (e.g., heavy cloud cover, geometric issues) or low scene complexity (e.g., extensive water bodies or snowfields), ensuring sufficient
LULC diversity and focus on areas of human activity. 
After filtering, 5709 flightlines retained. For each flight strip, metadata, including wavelength, full width at half maximum (FWHM), and ground sample distances (GSD) were extracted from acquisition records. Then, the selected AVIRIS HSI data were cropped into non-overlapping 256 $\times$ 256-pixel patches,
standardized to a uniform Float16 format for efficient SSL tasks. Finally, to ensure reliable input for MIM-based pre-training, 
patches located at the edges of the acquisition flightlines with over 20\% missing pixels were discarded. 
The final Hyper-400K dataset contains 400,000 high-quality patches, with a total volume of 17 TB, the detailed distribution of which is presented in Table~\ref{tab:Hyper-450K}.

\begin{table}[htbp]
\centering
\caption{Distribution of the Hyper-400K HSI dataset.}
\label{tab:Hyper-450K}
\footnotesize
\begin{tabular}{cccccccc}
\toprule
Sensor & \begin{tabular}[c]{@{}c@{}}Spectral\\range (nm)\end{tabular} & \begin{tabular}[c]{@{}c@{}}Spectral\\resolution (nm)\end{tabular} & \# Bands & \begin{tabular}[c]{@{}c@{}}Data\\level\end{tabular} & \# Flights & \begin{tabular}[c]{@{}c@{}}\# Image\\patches\end{tabular} & \begin{tabular}[c]{@{}c@{}}Spatial\\resolution (m)\end{tabular} \\ 
\hline
\multirow{2}{*}{AVIRIS-Classic} & \multirow{2}{*}{380\,\text{--}\,2500} & \multirow{2}{*}{10} & \multirow{2}{*}{224} & L1 & 556 & 39,046 & 0.6\,\text{--}\,18.0 \\
 & & & & L2 & 167 & 15,868 & 0.9\,\text{--}\,17.8 \\
\multirow{2}{*}{AVIRIS-NG} & \multirow{2}{*}{380\,\text{--}\,2510} & \multirow{2}{*}{5} & \multirow{2}{*}{425} & L1 & 2845 & 149,925 & 0.2\,\text{--}\,19.1 \\
 & & & & L2 & 1236 & 62,669 & 0.5\,\text{--}\,19.1 \\
\multirow{2}{*}{AVIRIS-3} & \multirow{2}{*}{390\,\text{--}\,2500} & \multirow{2}{*}{7.4} & \multirow{2}{*}{284} & L1 & 685 & 104,220 & 1.8\,\text{--}\,4.8 \\
 & & & & L2 & 220 & 29,266 & 1.2\,\text{--}\,4.4 \\
\bottomrule
\end{tabular}
\end{table}

\section{Proposed Methodology}
\subsection{Overview}

\begin{figure}[htb]
\vspace{-1.0em}
\centering
\includegraphics[width=0.95\linewidth]{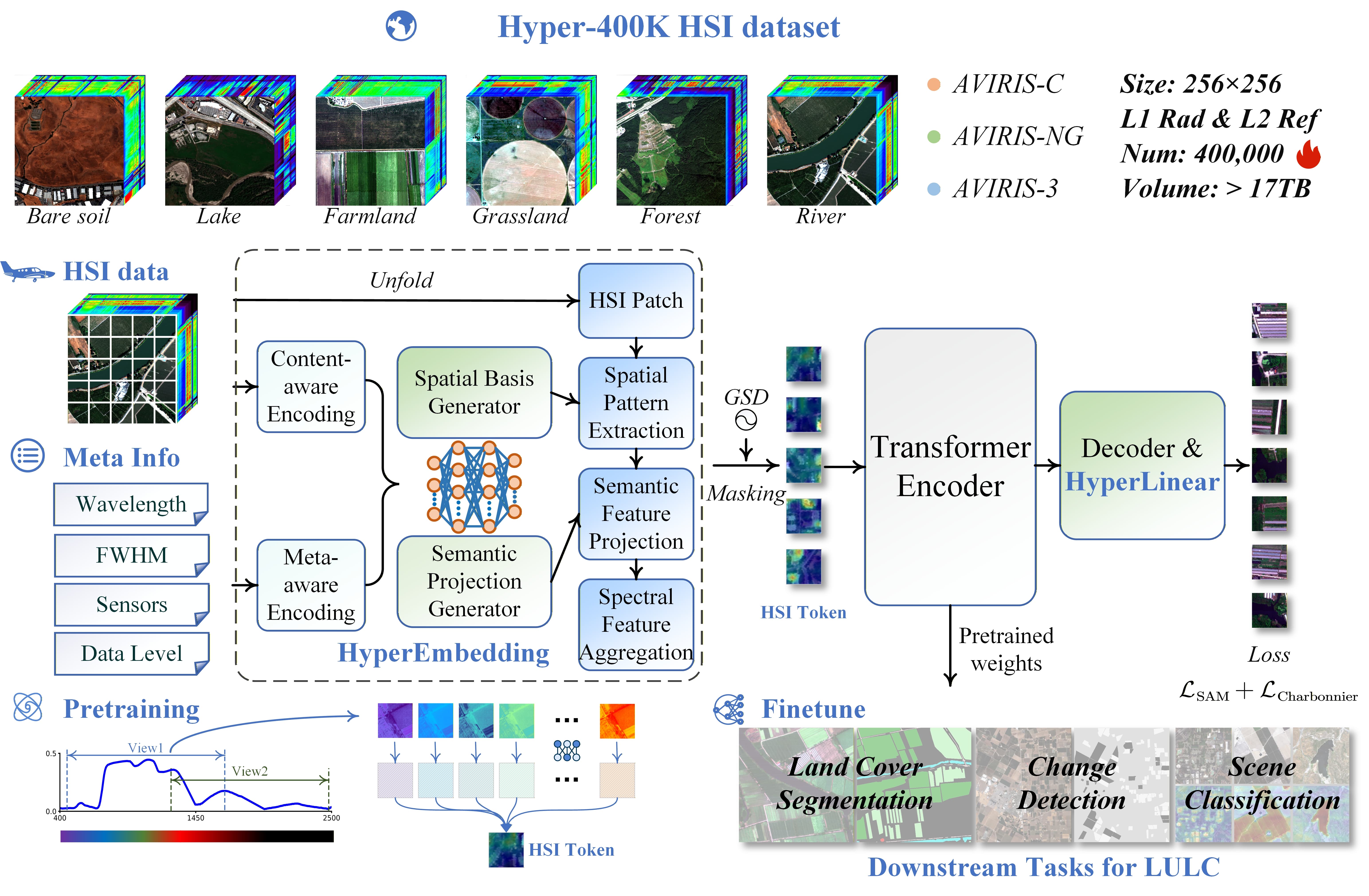}
\caption{Overview of the SpecAware pre-training framework. The framework starts with a large, diverse multi-sensor and multi-level aerial HSI dataset. The SpecAware model then uses a meta-content aware encoder and hypernetworks to perform channel-wise encoding of the spatial-spectral features, yielding a unified HSI token. This model was trained with a progressive scheme from a multi-view strategy and evaluated on three types of downstream tasks.}
\label{fig:main}
\end{figure}

In this work, a novel self-supervised learning framework for large-scale pre-training of HSI data, named SpecAware, was constructed. As shown in Figure~\ref{fig:main}, the SpecAware framework is built upon a condition-driven hypernetwork that integrates the principles of matrix decomposition, establishing a unified, end-to-end, masked self-supervised learning architecture for multi-sensor and multi-level HSI data. Firstly, SpecAware fuses textual and visual modalities by employing different attribute encoders and a spectral feature extractor to jointly encode the metadata and content features of the HSI data. 
The core mechanism of SpecAware is a sample-conditioned hypernetwork that dynamically generates two distinct matrix factors, a specific set of spatial and semantic matrix factors, for each training sample.
These two matrix factors adapt the representation layers to variations in metadata, spectral band count, and scene content, enabling sample-specific processing of spatial and spectral features for each band. Thirdly, we employ a progressive pre-training scheme that integrates both local and multi-view perspectives. Supported by a cloud-native, chunked data storage strategy, this approach makes efficient pre-training on massive HSI datasets feasible. The subsequent sections provide a comprehensive description of the proposed approach.

\subsection{Metadata- and content-aware encoding}

\begin{figure}[htbp]
\vspace{-1.0em}
\centering
\includegraphics[width=0.95\linewidth]{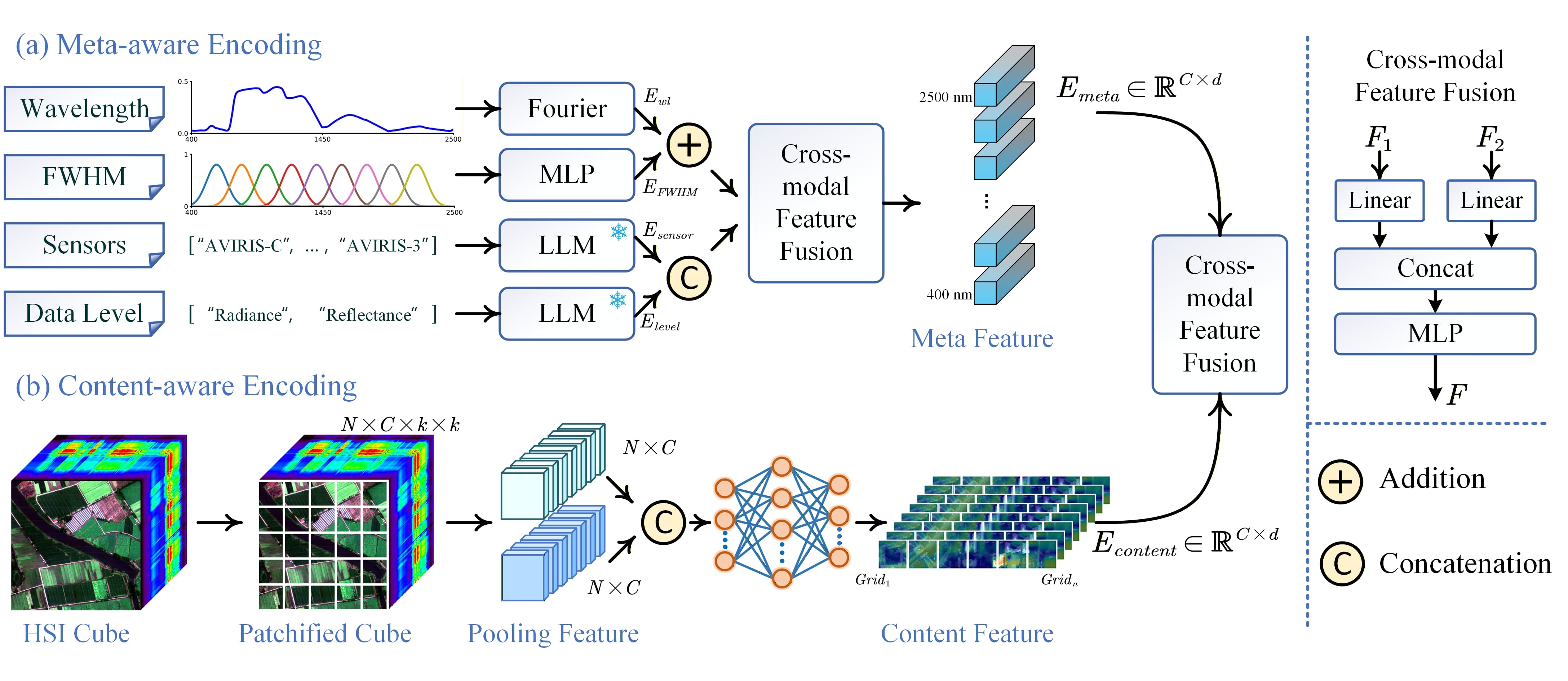}
\caption{(a) Sensor attributes meta-aware encoding module, and (b) image feature content-aware encoding module.}
\label{fig:encoding}
\vspace{-1.0em}
\end{figure}

As shown in Figure~\ref{fig:GA}, considering an HSI image $X$, the spectral feature distribution is shaped by the interplay between its meta-attributes (e.g., sensor type, data processing level) and its semantic content (e.g., the land-cover types depicted, such as trees, buildings, or water). To address this duality, and inspired by the Copernicus-FM model \citep{wangUnifiedCopernicusFoundation2025}, we propose a meta-feature joint encoding module, the architecture of which is detailed in Figure~\ref{fig:encoding}. This module fuses these metadata and content factors into a unified conditioning vector. This vector, in turn, drives the hypernetworks to dynamically generate instance-adaptive encoding weights tailored specifically for each input image from different HSI sensors with different data levels.

Firstly, HSI data are characterized by a large number of spectral bands $C$, and each associated with a physically meaningful wavelength. Thereby, inspired by the Aurora model \citep{bodnarFoundationModelEarth2025}, we apply Fourier encoding method to encode the wavelength of each band $\lambda\in\left\{\lambda_1,\ldots,\lambda_C\right\}$ into a $d$-dimensional vector $E_{wl} \in \mathbb{R}^{C \times d}$. For example, AVIRIS-3 sensor contains bands centered at wavelengths ranging from 390 nm to 2493 nm (e.g., 390, 397, …, 2493):
\begin{equation}
E_{wl} = FE(\lambda)
= \left[\, \cos\!\left(\frac{2\pi \lambda}{\omega_i}\right),\;  \sin\!\left(\frac{2\pi \lambda}{\omega_i}\right) \right],\ \forall\; 0 \le i < d/2.
\end{equation} where $\omega_i$ are log-spaced values between the minimum and maximum wavelength values:
\begin{equation}
\omega_i=\exp\left(\log\omega_{min}+i\cdot\frac{\log\omega_{max}-\log\omega_{min}}{d/2-1}\right).
\end{equation}

For most hyperspectral sensors, the spectral wavelength range is generally between 350 and 2550 nm. Therefore, we set the minimum value $\omega_{min}$ to 350 and the maximum value $\omega_{max}$ to 2550.

FWHM, denoted as $\delta\in\left\{\delta_1,\ldots,\delta_C\right\}$, is a complementary attribute to the center wavelength, characterizing the width of a sensor's spectral response, and thereby indicating its spectral resolution. Thus, we utilize a simple multilayer perceptron ($MLP$) to encode this critical attribute as $E_{FWHM}$.
For instance, the FWHM of AVIRIS-3 ranges from 8.30 nm to 7.62 nm across its spectral range:
\begin{equation}
  E_{FWHM}=MLP\left(\delta\right)\in \mathbb{R}^{C\times d}
\end{equation}

Then, for the encoding of textual data, such as sensor attributes and data processing level, and to ensure the future extensibility of the model, we employ a Large Language Model (LLM) encoding approach. The proposed method avoids the use of specialized encoders by using a frozen, pre-trained MiniLM model \citep{wangMinilmDeepSelfattention2020} to convert the sensor name (e.g.\ ``AVIRIS-C'', ``AVIRIS-NG'', and ``AVIRIS-3'') 
and data level (e.g., ``L1 Calibrated Radiance'' and ``L2 Surface Reflectance'')
into $d$-dimensional vectors. This one-time pre-processing step adds no overhead to the main pre-training, and a $Linear$ layer subsequently reduces the embedding dimension as $E_{name}$ and $E_{level}$. Since these attributes are constant across channels, they are broadcast along the spectral dimension as a unified sensor descriptor.
\begin{equation}
  E_{name}=Linear\left(LLM\left(Sensor\right)\right)\in \mathbb{R}^{C\times d/2}
\end{equation}
\begin{equation}
  E_{level}=Linear\left(LLM\left(Level\right)\right)\in \mathbb{R}^{C\times d/2}
\end{equation}

The above four attribute encodings can be categorized into two classes. Firstly, the spectral-physical features, which represent per-channel information, are generated by applying a weighted fusion with learnable scalars $\alpha$ and $\beta$ to the encodings of each band's center wavelength and FWHM. The fused representation is subsequently enhanced by an \textit{MLP} layer with residual connections. Secondly, the sensor-attribute features are formed by concatenating relevant attributes into a single vector.
\begin{equation}
  E_{spectral}=MLP\left(\alpha E_{wl}+\beta E_{FWHM}\right)
\end{equation}
\begin{equation}
  E_{sensor}=\mathrm{Concat}\left(E_{name},E_{level}\right)
\end{equation}

Then, to effectively fuse the spectral-physical features and sensor-attribute features, we designed a simple cross-modal feature fusion (CFF) module. Firstly, the two features are aligned using two independent linear projections and then concatenated. A residual \textit{MLP} subsequently processes the fused vector, reducing its dimensionality while modeling the complex interactions between sensor and spectral information.
\begin{equation}
  E_{spectral}^\prime=Linear\left(E_{spectral}\right)
\end{equation}
\begin{equation}
  E_{sensor}^\prime=Linear\left(E_{sensor}\right)
\end{equation}
\begin{equation}
  E_c=CFF\left(E_{spectral},E_{sensor}\right)=MLP\left(\mathrm{Concat}\left(E_{spectral}^\prime,E_{sensor}^\prime\right)\right)
\end{equation}

Finally, a lightweight spectral-aware Transformer applies self-attention to the band embedding for contextual enhancement, producing the final enhanced meta-embedding $E_{meta}$:
\begin{equation}
  E_{meta}=Transformer\left(E_c\right)\in \mathbb{R}^{C\times d}
\end{equation}

On the other hand, to capture the intrinsic spectral and spatial characteristics of HSI image $X\in\mathbb{R}^{C\times H\times W}$, where $C$, $H$, and $W$ are the number of channels, spatial height, and width, respectively, we introduce a dual-pooling strategy to extract patch-level scene features. Specifically, both an average pooling and a max pooling layer are applied to the feature map $X$ with a non-overlapping kernel of size $k\times k$, which is identical to the patch size of the MAE encoder. This process yields two sets of features, $X_{\mathrm{avg}}$ and $X_{\mathrm{max}}$, representing the average and maximum values within each patch-sized region, respectively.
\begin{equation}
  X_{\mathrm{avg}}={\mathrm{AvgPool2d}}_k\left(X\right)\in \mathbb{R}^{C\times N},\ \ X_{\mathrm{max}}={\mathrm{MaxPool2d}}_k\left(X\right)\in \mathbb{R}^{C\times N}
\end{equation} where $N = (H/k) \times (W/k)$ is the number of non-overlapping patches. 
Subsequently, the two pooled features are then concatenated and mapped to a lower-dimensional space using stacked \textit{MLP} layers to generate the final content features $E_{content}$, which can be defined as:
\begin{equation}
  E_{content}=MLP\left(\mathrm{Concat}\left(X_{\mathrm{avg}},\ X_{\mathrm{max}}\right)\right)\in \mathbb{R}^{C\times d}
\end{equation}

Finally, the meta features $E_{meta}$ and content features $E_{content}$ are fused. Given that this is also a cross-modal fusion task, we reuse the proposed simple CFF module to create a comprehensive conditional vector \textit{E}, enabling dynamic content awareness for each HSI cube, which in turn drives the two subsequent hypernetworks.

\begin{equation}
  E=CFF\left(E_{meta},E_{content}\right)\in \mathbb{R}^{C\times d}
\end{equation}

Notably, the encoding process dynamically generates an instance-specific and content-adaptive embedding for each sample in one training batch. For simplicity, the batch dimension \textit{B} has been omitted in the formulas.

\subsection{Hypernetwork-driven dynamic spatial-spectral embedding}

\begin{figure}[tbp]
\vspace{-1.0em}
\centering
\includegraphics[width=0.95\linewidth]{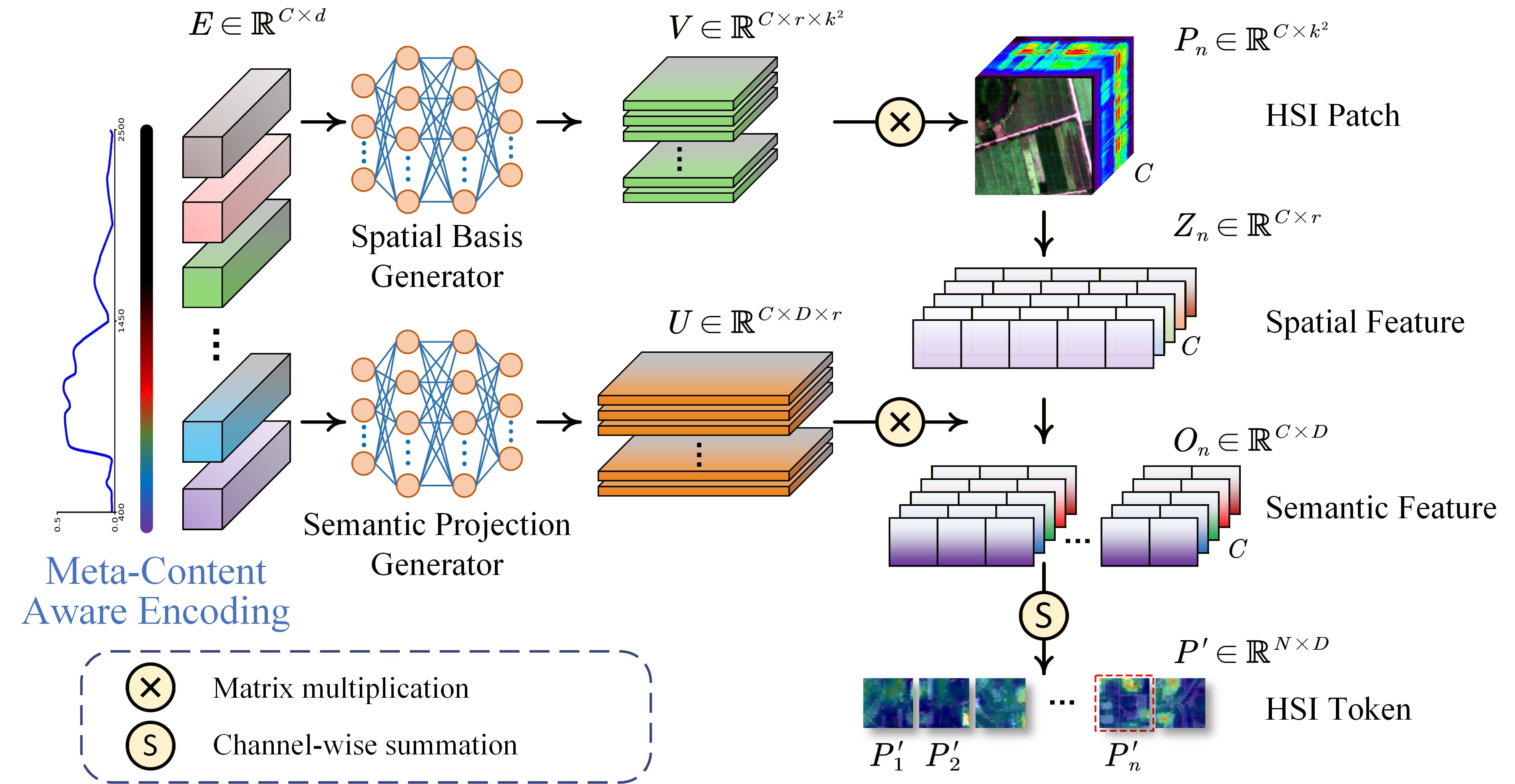}
\caption{Architecture of the HyperEmbedding module. The module employs hypernetworks that take fused features as input to dynamically generate matrix parameters for adaptive spatial pattern extraction and latent semantic feature projection.}
\label{fig:matrix}
\vspace{-1.0em}
\end{figure}

To overcome the limitations of static weights in handling variable-channel HSI data, we introduce a sample-level conditional hypernetwork framework for dynamic spatial-spectral embedding, named HyperEmbedding. Inspired by the intrinsic low-rank property of HSI data and matrix decomposition theory, the proposed approach diverges from generating high-dimensional convolution kernels for variable-channel data. Instead, a hypernetwork, conditioned on both sensor meta and image content features, dynamically generates a pair of lightweight matrix factors for each sample on a per-channel basis. As illustrated in Figure~\ref{fig:matrix}, this matrix factorization decouples the transformation of HyperEmbedding into three stages—adaptive spatial pattern extraction, latent semantic feature projection, and spectral feature aggregation. The per-channel design gives the HyperEmbedding module inherent flexibility to handle variations in the number of spectral channels (e.g., from a new sensor or after removing noisy bands), as it can adaptively add or remove the corresponding channel-wise computations without altering the model architecture.

We follow the processing paradigm of MAE to perform an unfold operation on the input ${X}$ to extract non-overlapping patches of size $k\times k$. Following tensorization and dimension rearrangement, the original HSI image is converted into a sequence of patch tokens, represented by a tensor $P\in\mathbb{R}^{N\times C\times k^2}$. Here, $N=\left(H/k\right)\times\left(W/k\right)$ is the total number of patches, and $k^2$ represents the flattened spatial dimension of each patch.

We next detail the construction of hypernetworks, consisting of the spatial basis and semantic projection generator. Their core function is to take the conditional vector \textit{E}, which fuses image metadata with content features, into the decomposition parameters required by the main network. Three hypernetworks consist of stacked fully connected layers and non-linear activation functions, and they respectively generate the spatial decomposition matrix \textit{V}, the latent feature projection matrix \textit{U}, and a globally adaptive bias vector $bias$. Thereby, for each individual sample in the batch, the proposed hypernetwork architecture instantiates instance-specific transformation parameters for each of its spectral channels, enabling the sample-level dynamic behavior.

\begin{equation}
  U=\mathrm{Reshape}\left(\mathcal{F}_U\left(E\right)\right)\in\mathbb{R}^{C\times D\times r}
\end{equation}
\begin{equation}
  V=\mathrm{Reshape}\left(\mathcal{F}_V\left(E\right)\right)\in\mathbb{R}^{C\times r\times k^2}
\end{equation}
\begin{equation}
  bias=\mathcal{F}_b\left(\mathrm{mean}\left(E\right)\right)\in\mathbb{R}^D
\end{equation}where $\mathcal{F}_U$, $\mathcal{F}_V$, and $\mathcal{F}_b$ denote the hypernetwork module, and \textit{D} is the target feature dimension, set to match the input requirement of the subsequent ViT backbone.

Next, we utilize the decomposition matrices \textit{V} and \textit{U} to apply a two-stage adaptive linear transformation to each HSI patch. The first stage is adaptive spatial pattern extraction. The hypernetwork-generated matrix \textit{V} acts as an adaptive spatial feature extractor, which linearly remaps each original $k^2$-dimensional patch vector $P_{n}\in\mathbb{R}^{C\times k^2}$ into a $r$-dimensional latent space $Z_{n}\in\mathbb{R}^{C\times r}$. This process is designed to capture the most informative spatial structures and textural patterns, and can be formulated as:

\begin{equation}
  Z_{n,c}=P_{n,c}\cdot V_c^\top
\end{equation}where $Z\in\mathbb{R}^{N\times C\times r}$ is the latent space representation of HSI patch \textit{P}, and \textit{r} is the dimension of the latent space.

The second stage is latent semantic feature projection. The matrix \textit{U} is responsible for encoding and mapping the $r$-dimensional latent feature $Z_{n}$ into a more discriminative and semantically rich feature representation within the final $D$-dimensional embedding space. 

\begin{equation}
  O_{n,c}=Z_{n,c}\cdot U_c^\top
\end{equation}where $O\in\mathbb{R}^{N\times C\times D}$ is the embedding contribution for each channel \textit{c} of HSI patch \textit{P}.

Finally, to form the unified, channel-agnostic embedding representation $HyperEmbedding\in\mathbb{R}^{N\times D}$ for the HSI patch, we perform spectral feature aggregation on the per-channel contribution vectors, applying a summation along the channel dimension \textit{C}. Thus, the entire channel-dynamic HyperEmbedding process can be holistically represented by the following equation:
\begin{equation}
  {\mathrm{HyperEmbedding}}_n=\sum_{c=1}^{C}O_{n,c}=\sum_{c=1}^{C}\left(\left(P_{n,c}\cdot V_c^\top\right)\cdot U_c^\top\right)+bias
\end{equation}

Through the HyperEmbedding module, driven by the hypernetworks and matrix factorization, the SpecAware method can be adapted to any HSI data with a different input band count.

\subsection{Hypernetwork-driven dynamic decoder for HSI data}

Following the same principle as the encoder, we designed a dynamic hypernetwork-based linear layer, named HyperLinear, to replace the traditional static linear layer, which is conditioned on meta- and content-aware features to adapt to varying HSI channel counts. Notably, the meta-aware encoding used here is consistent with that in the HyperEmbedding stage. 

The content-aware features are extracted directly from the deep latent features $X^\prime\in\mathbb{R}^{N\times D}$ output by the MAE decoder. Specifically, we first distill a global content representation $F^\prime\in\mathbb{R}^{1\times D}$ using global average pooling and a lightweight Transformer module. Then, it is broadcast to a shape of $\mathbb{R}^{C\times D}$ to align with the meta info features for subsequent cross-modal fusion.

Similarly to the HyperEmbedding module, the meta-content fused features $E^\prime$ drive the hypernetwork to generate feature reconstruction matrices 
$U^\prime \in \mathbb{R}^{C \times k^{2} \times r }$, 
$V^\prime \in \mathbb{R}^{C \times r \times D}$ 
and 
$bias^\prime \in \mathbb{R}^{C \times k^{2}}$
, which enables flexible and adaptive decoding for different HSI data. For each latent vector $x_n\in\mathbb{R}^D$, we first replicate it along $C$ channels to enable $x_n^\prime\in\mathbb{R}^{C\times D}$, then $x_n^\prime$ is combined with the generated reconstruction matrices to project back into the full HSI patch space.
\begin{equation}
  {HyperLinear}_n=x_n^\prime\cdot V^{\prime \top}\cdot U^{\prime \top}+{bias}^\prime
\end{equation}

\subsection{ViT encoder and decoder}
Following the standard MAE training paradigm \citep{heMaskedAutoencodersAre2022}, we adopt the vanilla ViT encoder-decoder architecture and random masking strategy. However, considering that airborne hyperspectral data exhibit substantial variation spatial resolution, from 0.2 m to 19.1 m within the Hyper-400K dataset, we replace the standard fixed positional encoding with a GSD-aware positional encoding proposed in Scale-MAE \citep{reedScalemaeScaleawareMasked2023}.

\subsection{Reconstruction loss function}

Despite a meticulous preprocessing pipeline for the HSI data is applied, including outlier removal and band-wise normalization, inherent noise in such high-dimensional imagery remains a challenge. The mean squared error (MSE) loss is highly sensitive to this noise due to its quadratic penalty mechanism, which can compromise both convergence and generalization. Therefore, we introduce a hybrid loss function that synergistically combines the Charbonnier and spectral angle mapper (SAM) losses. The Charbonnier loss $\mathcal{L}_{\mathrm{Charbonnier}}$, as a robust variant of the L1 loss \citep{barronGeneralAdaptiveRobust2019}, dampens the impact of large-magnitude errors from noise. Concurrently, the SAM loss $\mathcal{L}_{SAM}$ \citep{tu2HM22024} focuses on preserving spectral fidelity by minimizing the angular deviation between reconstructed and original vectors, thus learning features that are invariant to illumination changes. In general, the network is then trained by minimizing the total loss function $\mathcal{L}_{total}$, which can be defined as:

\begin{equation}
  \mathcal{L}_{\mathrm{Charbonnier}}=\sqrt{(x-\hat{x})^2+\epsilon^2}
\end{equation}
\begin{equation}
  \mathcal{L}_{SAM}=1-\mathrm{CosSim}\left(x,\hat{x}\right)
\end{equation}
\begin{equation}
  \mathcal{L}_{total}=\alpha\mathcal{L}_{\mathrm{Charbonnier}}+\beta\mathcal{L}_{SAM}
\end{equation}where $x$ and $\hat{x}$ represent the original and reconstructed spectral vectors, respectively, the term $\epsilon$ is a small positive constant introduced to ensure differentiability, and $\mathrm{CosSim}(\cdot)$ denotes the cosine similarity function. In the experiments conducted in this study, we set the loss weights $\alpha$ and $\beta$ to 2 and 1, respectively, to balance the two loss items.
\subsection{Progressive pre-training}

Inspired by SpectralGPT \citep{hongSpectralGPTSpectralRemote2024} and HyperSL \citep{kongHyperSLSpectralFoundation2025}, we adopted a progressive pre-training strategy that gradually increases the diversity and complexity of training data across three stages. The model is first pretrained on the high-quality AVIRIS-3 L1 dataset to capture fundamental spectral characteristics under a single-sensor setting. It is then exposed to additional sensor types and processing levels to capture inter-sensor differences and domain variability (e.g. radiance and reflectance). For this stage, a 90,000-image subset (Hyper-90K) is constructed by sampling 20,000 images from each of the three L1 datasets and 10,000 images from each of the three L2 datasets. Finally, pretraining on the full Hyper-400K dataset enables the model to learn broad spatial coverage, resulting in more generalizable feature representations.

\section{Experiments and Analysis}

\subsection{Experimental settings}

Self-supervised pretraining was conducted on the constructed Hyper-400K dataset, stored in Zarr format \citep{gowanUsingCloudComputing2022} with FP16 precision to improve data loading efficiency. All the pretraining experiments were conducted on eight NVIDIA RTX 4090 GPUs using the PyTorch 2.5 framework, with TorchData enabling mixed-sensor training. The proposed matrix factorization operations were implemented with GPU parallelized einsum to ensure the computational efficiency \citep{blacherEinsumBenchmarkEnabling2024}. 

We utilized the ViT-Base model as the backbone, with an input patch size of $8\times8$, and followed the MAE pre-training framework. Input images were randomly cropped to $224\times224$ and normalized using 1\% percentile clipping and the sensor-specific mean and standard deviation. To ensure compatibility with new, unknown sensors, a 10\% random dropout was applied to the sensor names in the input meta-attributes during training. We set the image masking ratio to 0.75, using the AdamW optimizer with an effective batch size of 256. 

\subsection{Pretraining strategy}

To efficiently process large-scale HSI data with limited computational resources, we employed a multi-view pre-training strategy integrated with distributed parallel training. This strategy employs random continuous band sampling, selecting 100 continuous bands with a stride of 32, in a batch level for each GPU. In a distributed training environment, the strategy generates distinct spectral views for different image batches on different GPU devices, which effectively enhances model robustness while balancing computational and data loading efficiency. 

The model was first pretrained for 800 epochs on the AVIRIS-3 L1 dataset (approximately 30 hours). Next, it was trained for an additional 600 epochs (approximately 17.5 hours) on the Hyper-90K subset. Finally, the model was trained for 200 epochs on the complete Hyper-400K dataset (approximately 27.5 hours). The multi-view pre-training approach was used for all three pre-training stages. For each stage, the initial learning rate was set to 0.00015, decaying to a minimum of 0.00001. The warmup periods were set to 20 and 10 epochs for the first two stages and the final stage, respectively.

\subsection{Hyperspectral land-cover semantic segmentation}

LULC classification constitutes a classic task in the interpretation of HSI data. The increasing availability of large-scale, high-resolution airborne HSI data has rendered pixel-wise classification methods inefficient for LULC interpretation tasks. Consequently, we conducted end-to-end patch-free land-cover classification at a semantic segmentation level on the three large-scale airborne HSI datasets.

\subsubsection{Dataset description}

To evaluate the performance of the SpecAware model in HSI LULC classification downstream tasks, we conducted semantic segmentation experiments on three large-scale aerial hyperspectral datasets, namely, AeroRIT \citep{rangnekarAeroRITNewScene2020}, Qingpu-HSI \citep{jiPatchOutNovelPatchfree2025}, and WHU-H$^2$SR \citep{tu2HM22024}.

1) The AeroRIT dataset was acquired by a Headwall sensor over the campus of Rochester Institute of Technology, USA \citep{rangnekarAeroRITNewScene2020}. The original data comprise $1973\times3975$ spatial pixels at a 0.4-m resolution, 372 bands spanning 397--1003\,nm, and five labeled land-cover classes. We used the surface reflectance version and, following \cite{rangnekarAeroRITNewScene2020}, cropped it to a 305-band subset spanning $\sim$400--897\,nm. We cropped the dataset into non-overlapping 128 $\times$ 128-pixel patches, which were then split into training (20\%), validation (10\%), and test (70\%) sets.

2) The Qingpu-HSI dataset was captured in Qingpu District, Shanghai, China, using the AMMIS sensor \citep{jiPatchOutNovelPatchfree2025}. It comprises 251 spectral bands with $20480\times2944$ pixels at a 0.75-m spatial resolution and features manual annotations for 20 distinct land-cover classes. Due to the dataset's sparse annotations, we followed the training portioning of PatchOut \citep{jiPatchOutNovelPatchfree2025} and augmented the data with the resampling method from FreeNet \citep{zhengFPGAFastPatchFree2020}, sampling 2000 pixels from each land-cover class from 16 training areas. The classification accuracy was evaluated on the test partitions of the scene, after generating a complete prediction map for the entire scene using an overlapping patch strategy.

3) The WHU-H$^{2}$SR dataset, which is also from the AMMIS sensor, covers a 227.79 km$^{2}$ area in southern Shenyang, China \citep{tu2HM22024, yuSTSNetCrossspatialResolution2025}. It is composed of 2531 sub-images, each of $300\times300$ pixels, with a 1-m spatial resolution. The dataset provides surface reflectance data across 249 spectral bands. For the experiments, it was divided into training, validation, and test sets, comprising 20\%, 10\%, and 70\% of the data, respectively. For the training, we randomly cropped $256\times256$ patches from the images. For the testing, we applied a consistent center crop of the same dimensions.

\subsubsection{Experimental settings}

\begin{figure}[tbp]
\vspace{-1.0em}
\centering
\includegraphics[width=0.75\linewidth]{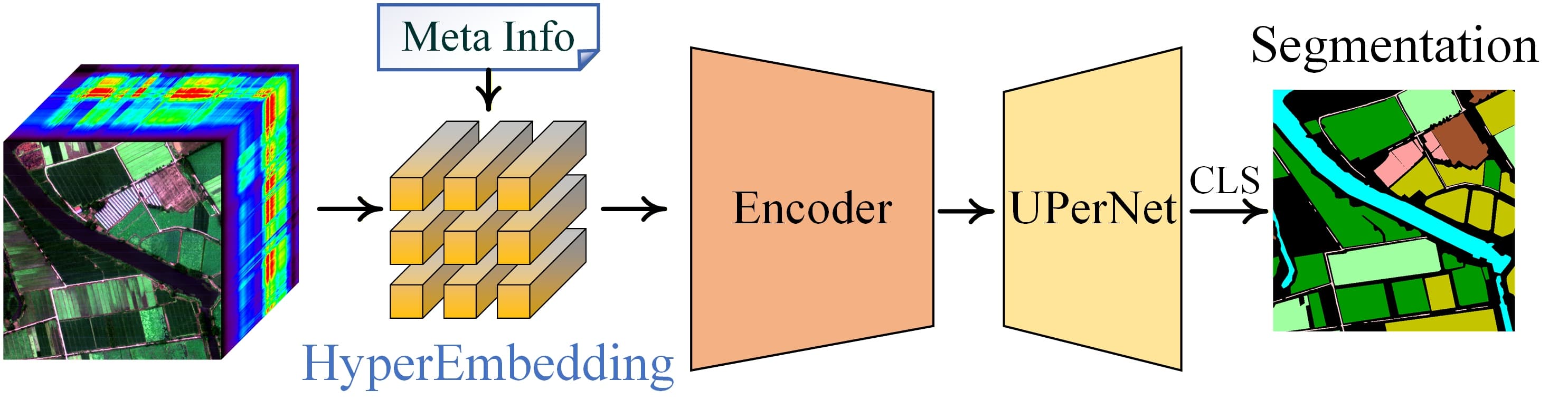}
\caption{The fine-tuning framework of SpecAware for the semantic segmentation task.}
\label{fig:Seg}
\vspace{-1.0em}
\end{figure}

Following the fine-tuning paradigm illustrated in Figure~\ref{fig:Seg}, for SpecAware, we employed the pre-trained encoder and appended a UPerNet decoder \citep{xiaoUnifiedPerceptualParsing2018} for the semantic segmentation task. For the AeroRIT dataset, an input size of $128\times128$ and a token patch size of 4 were used. Separately, for the Qingpu-HSI and WHU-H$^{2}$SR datasets, an input size of $256\times256$ and a token patch size of 8 were adopted. We trained the model for 100 epochs on all three datasets, using a weighted cross-entropy loss function, with a batch size of 16 and an initial learning rate of 0.00005.

To comprehensively evaluate the performance of the proposed SpecAware model, we benchmarked the model against the supervised models of ABCNet \citep{liABCNetAttentiveBilateral2021}, FreeNet \citep{zhengFPGAFastPatchFree2020}, UNetFormer \citep{wangUNetFormerUNetlikeTransformer2022}, UM2Former \citep{xuUM2FormerUShapedMultimixed2025}, and PatchOut \citep{jiPatchOutNovelPatchfree2025}, as well as the pre-trained models of SpectralEarth \citep{brahamSpectralEarthTrainingHyperspectral2025} and HyperSIGMA \citep{wangHypersigmaHyperspectralIntelligence2025}. 
For SpectralEarth, we also employed the UPerNet decoder, whereas for HyperSIGMA, we adhered to its official implementation, retaining the original spatial-spectral dual-branch architecture.
Model performance is assessed using the overall accuracy (OA), average accuracy (AA), Kappa, and mean intersection over union (mIoU).

\subsubsection{Results and analysis}

Tables~\ref{tab:Seg_AeroRIT},~\ref{tab:Seg_Qingpu} and~\ref{tab:Seg_H2SR} summarize the quantitative performance of the various models on the three semantic segmentation datasets. Overall, the proposed SpecAware model demonstrates the superior performance among all the compared methods. It achieves an OA of 92.85\%, 96.68\%, and 89.72\% and an mIoU score of 78.78\%, 59.57\%, and 67.44\% on the AeroRIT, Qingpu-HSI, and WHU-H$^2$SR datasets, respectively, which represents robust generalization capability to the airborne LULC classification tasks.

Specifically, as shown in Table~\ref{tab:Seg_AeroRIT}, on the AeroRIT dataset, the proposed SpecAware model achieves the superior performance, which obtains scores that are 0.75\% to 3.56\% higher in OA and 0.94\% to 10.02\% in mIoU, compared with other methods. These results indicate the effective fine-tuned learning capability of SpecAware on the AeroRIT dataset, which has a very high spatial resolution of 0.4\,m for HSI data, particularly for tasks requiring spatial localization and interpretation.

\begin{figure}[!th]
\centering
\includegraphics[width=0.98\linewidth]{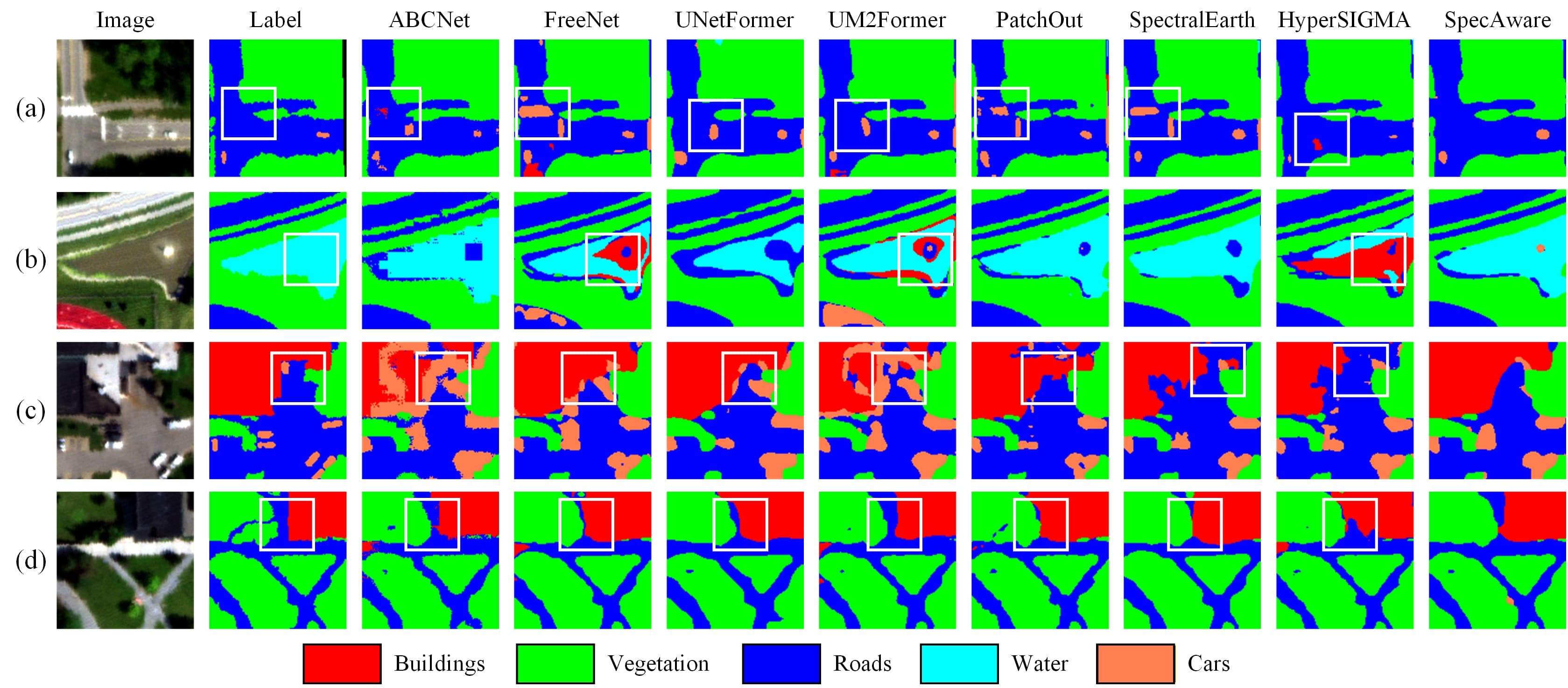}
\caption{Visual comparison of the HSI data land-cover segmentation results of the different methods on the AeroRIT dataset.}
\label{fig:Seg_res1}
\vspace{-0.8em}
\end{figure}

Figure~\ref{fig:Seg_res1} presents a visual comparison of the different methods on the AeroRIT dataset. For scene (a), the FreeNet, UNetFormer, and SpectralEarth models misclassify road markings as cars. In scene (b), the HyperSIGMA and UM2Former models misidentify the water body as buildings. In scene (c), ABCNet and UM2Former exhibit the confusion between buildings and roads. Finally, scene (d) demonstrates a common limitation, where the presence of strong shadows causes most of the methods to mislabel shadowed roads as buildings.

\begin{table}[!th]
\caption{Quantitative results of the different classification methods on the AeroRIT dataset.}
\label{tab:Seg_AeroRIT}
\scriptsize
\centering
\makebox[\textwidth][c]{
  \begin{tabular}{c *{8}{C{1.8cm}}} 
  \toprule
  Class & ABCNet & FreeNet & UNetFormer & UM2Former & PatchOut & SpectralEarth & HyperSIGMA & SpecAware \\ 
  \hline

C1 & \uline{88.32} & 87.26 & 78.92 & 81.69 & 83.56 & 81.97 & 82.23 & \textbf{91.91} \\
C2 & 94.89 & \textbf{97.29} & 94.87 & 96.11 & \uline{96.93} & 96.59 & 96.80 & 96.41 \\
C3 & 82.37 & 79.05 & 86.98 & 82.72 & 87.21 & \textbf{89.78} & 86.76 & \uline{87.62} \\
C4 & 89.36 & 82.87 & 82.90 & 79.45 & 84.96 & \textbf{96.69} & 84.79 & \uline{90.37} \\
C5 & 79.08 & 78.93 & 80.04 & 80.12 & 75.96 & \uline{82.24} & 67.39 & \textbf{91.93} \\
OA & 89.64 & 89.57 & 89.65 & 89.29 & 91.38 & \uline{92.10} & 90.80 & \textbf{92.85} \\
AA & 86.80 & 85.08 & 84.74 & 84.02 & 85.72 & \uline{89.45} & 83.60 & \textbf{91.65} \\
Kappa & 83.56 & 83.47 & 83.49 & 82.99 & 86.21 & \uline{87.35} & 85.22 & \textbf{88.63} \\
mIoU & 71.70 & 68.76 & 71.61 & 69.34 & 75.13 & \uline{77.84} & 73.73 & \textbf{78.78} \\

  \bottomrule
  \multicolumn{9}{l}{* Class labels correspond to: C1 Buildings, C2 Vegetation, C3 Roads, C4 Water, C5 Cars.} \\
  \end{tabular}
}
\end{table}

We further assessed the model on the Qingpu-HSI dataset, which features sparse labels and significant class imbalance. In this evaluation, the proposed SpecAware model achieves the superior performance, with an OA of 96.68\% and an mIoU of 59.57\%. These results demonstrate the model's generalization capability, suggesting that it can maintain high performance with relatively limited labeled data

\begin{figure}[tbp]
\vspace{-1.0em}
\centering
\includegraphics[width=0.98\linewidth]{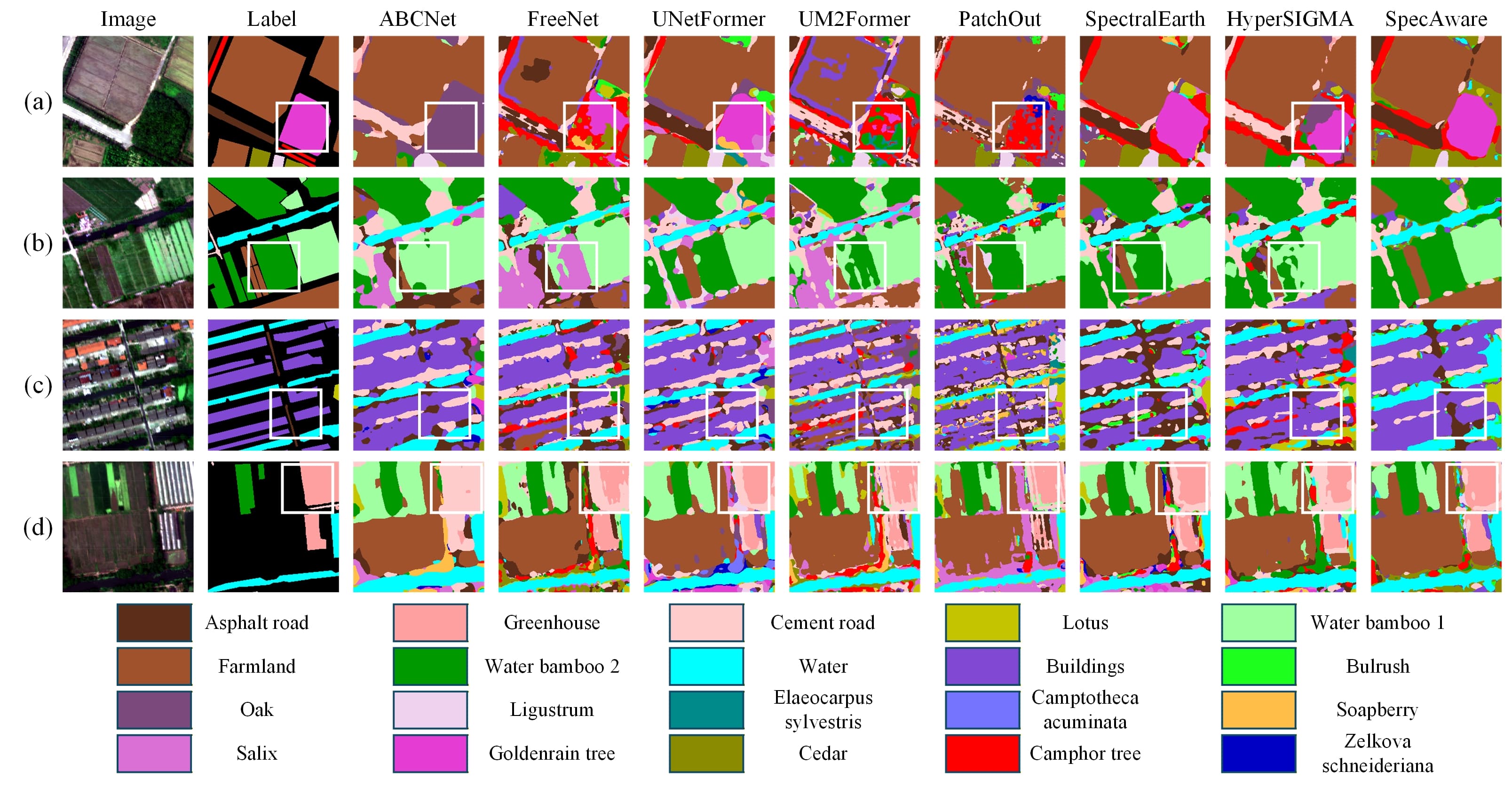}
\caption{Visual comparison of the HSI data land-cover segmentation results of the different methods on the Qingpu-HSI dataset.}
\label{fig:Seg_res2}
\vspace{-1.0em}
\end{figure}

As shown in Figure~\ref{fig:Seg_res2}, scene (a) is a tree species classification task, which shows that SpecAware, UNetFormer, and SpectralEarth perform well on the Goldenrain tree class. Scene (b) contains two varieties of water bamboo at different growth stages, where the predictions from ABCNet, UM2Former, and HyperSIGMA are fragmented and inconsistent. In the rural residential example (c), the segmentation of SpecAware is the most spatially coherent, whereas some models confuse asphalt road with building. In the greenhouse scene (d), the ABCNet, FreeNet, and HyperSIGMA models mislabel the greenhouse as cement road.

\begin{table}[htbp]
\caption{Quantitative results of the different classification methods on the Qingpu-HSI dataset.}
\label{tab:Seg_Qingpu}
\scriptsize 
\centering 
\makebox[\textwidth][c]{
  \begin{tabular}{c *{8}{C{1.8cm}}} 
\toprule
Class & ABCNet & FreeNet & UNetFormer & UM2Former & PatchOut & SpectralEarth & HyperSIGMA & SpecAware \\ 
\hline

C1  & 70.42 & 61.22 & 71.35 & 70.90 & 75.13 & \uline{82.33} & 71.75 & \textbf{84.28} \\
C2  & 85.92 & \uline{93.72} & \textbf{94.86} & 89.81 & 91.01 & 91.91 & 91.80 & 93.54 \\
C3  & 81.99 & 89.30 & 91.81 & \textbf{94.99} & 94.11 & \uline{94.49} & 82.44 & 93.30 \\
C4  & 95.17 & 92.84 & 92.05 & 92.84 & \uline{96.59} & 95.53 & 95.82 & \textbf{98.11} \\
C5  & 90.70 & 92.44 & \uline{94.11} & 91.61 & 89.20 & \textbf{95.41} & 91.51 & 93.34 \\
C6  & 93.02 & 88.43 & 96.50 & 89.46 & \uline{97.22} & 96.99 & 94.69 & \textbf{97.68} \\
C7  & 87.35 & 82.84 & 87.42 & 85.06 & 92.50 & \textbf{97.75} & 86.12 & \uline{95.59} \\
C8 & 98.33 & \uline{98.68} & 98.36 & 98.02 & 98.33 & 98.30 & \textbf{98.94} & 98.23 \\
C9  & 83.00 & 90.95 & 85.00 & 79.80 & 83.61 & 90.32 & \uline{93.94} & \textbf{96.61} \\
C10 & 62.87 & 60.63 & 60.92 & 60.54 & 65.83 & \uline{74.32} & 63.90 & \textbf{76.89} \\
C11 & \textbf{100.00} & 99.26 & \uline{99.95} & \uline{99.95} & \textbf{100.00} & 97.72 & 98.04 & 96.50 \\
C12 & 27.51 & 17.09 & 20.43 & 18.71 & 27.22 & \uline{27.53} & 19.08 & \textbf{30.87} \\
C13 & 44.88 & 79.09 & 73.70 & 77.72 & 81.27 & \textbf{92.28} & 76.90 & \uline{91.25} \\
C14 & \uline{95.86} & 69.25 & \textbf{99.00} & 80.62 & 95.43 & 86.15 & 78.61 & 85.32 \\
C15 & \textbf{64.33} & 39.25 & 47.60 & 47.40 & 41.84 & 44.95 & 41.41 & \uline{57.46} \\
C16 & 14.03 & 5.25 & \textbf{67.77} & 2.34 & 28.20 & \uline{33.36} & 9.43 & 30.93 \\
C17 & 81.03 & 69.19 & 85.84 & 45.92 & 49.20 & \textbf{95.24} & 79.36 & \uline{85.97} \\
C18 & 48.91 & 44.03 & 53.20 & 31.29 & \uline{63.31} & 50.72 & 43.33 & \textbf{65.97} \\
C19 & 19.33 & 52.72 & 55.87 & 43.83 & 67.60 & \uline{73.82} & 70.73 & \textbf{74.80} \\
C20 & \textbf{67.54} & 4.74 & 35.40 & 4.35 & 38.37 & 39.45 & \uline{67.06} & 17.96 \\
OA  & 92.86 & 92.49 & 94.58 & 91.90 & 95.12 & \uline{96.25} & 94.86 & \textbf{96.68} \\
AA  & 70.61 & 66.55 & 75.56 & 65.26 & 73.80 & \uline{77.93} & 72.74 & \textbf{78.23} \\
Kappa & 89.80 & 89.28 & 92.24 & 88.45 & 93.00 & \uline{94.62} & 92.63 & \textbf{95.23} \\
mIoU  & 43.28 & 43.89 & 47.90 & 42.03 & 51.19 & \uline{57.96} & 51.33 & \textbf{59.57} \\

  \bottomrule
  \multicolumn{9}{p{1.1\textwidth}}{* Class labels correspond to: C1 Asphalt road, C2 Greenhouse, C3 Cement road, C4 Lotus, C5 Water bamboo 1, C6 Farmland, C7 Water bamboo 2, C8 Water, C9 Buildings, C10 Bulrush, C11 Oak, C12 Ligustrum, C 13 Elaeocarpus sylvestris, C14 Camptotheca acuminata, C15 Soapberry, C16 Salix, C17 Goldenrain tree, C18 Cedar, C19 Camphor tree, C20 Zelkova schneideriana.} \\
  \end{tabular}
}
\end{table}

The WHU-H$^2$SR dataset is the largest of the three LULC classification benchmarks in both scale and geographic coverage, focusing primarily on rural scenes. On this dataset, the proposed SpecAware model achieves the highest OA of 89.72\% and mIoU of 67.44\%, which obtains scores that are 2.12\% to 7.71\% higher in OA and 4.52\% to 13.59\% in mIoU, compared with other methods. For a semantic segmentation task, the higher mIoU score suggests that the SpecAware model has a more balanced performance across all the land-cover classes.

\begin{figure}[!tb]
\centering
\includegraphics[width=0.98\linewidth]{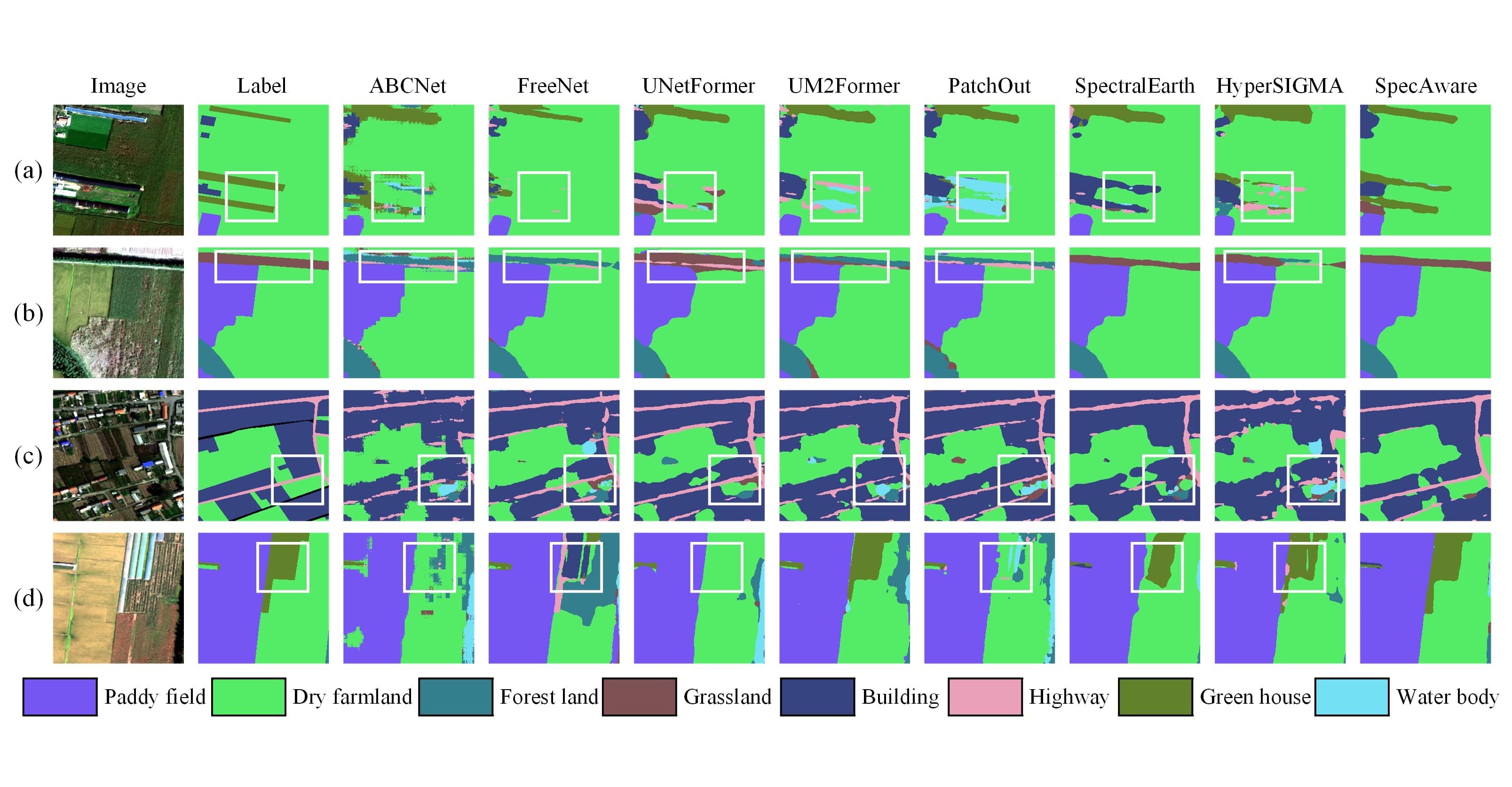}
\caption{Visual comparison of the HSI data land-cover segmentation results of the different methods on the WHU-H$^2$SR dataset.}
\label{fig:Seg_res3}
\end{figure}

Figure \ref{fig:Seg_res3} displays the qualitative semantic segmentation results of the various models on the WHU-H$^2$SR dataset. The results again illustrate the complexity of the greenhouse class, particularly those with varying colors. In scenes (a) and (d), while many of the models struggle, the proposed SpecAware model provides a more complete segmentation of the greenhouse structures. In scene (b), several models, e.g., ABCNet and FreeNet, confuse grassland with other classes. Although HyperSIGMA correctly classifies the type, its segmentation is fragmented. Scene (c) demonstrates the impact of shadows, causing some of the models to misclassify dry farmland as buildings or water. In general, SpecAware again generates the most intact and accurate results.

\begin{table}[tbp]
\caption{Quantitative results of the different classification methods on the WHU-H$^2$SR dataset.}
\label{tab:Seg_H2SR}
\scriptsize 
\centering 
\makebox[\textwidth][c]{
  \begin{tabular}{c *{8}{C{1.8cm}}} 
\toprule
Class & ABCNet & FreeNet & UNetFormer & UM2Former & PatchOut & SpectralEarth & HyperSIGMA & SpecAware \\ 
\hline

C1 & 93.49 & 94.48 & 94.90 & 94.24 & 95.76 & \uline{96.71} & 95.05 & \textbf{96.94} \\
C2 & 79.59 & 84.97 & 83.93 & 83.63 & 85.60 & \uline{87.34} & 86.66 & \textbf{91.08} \\
C3 & 69.65 & 67.50 & \uline{76.22} & 69.70 & 71.85 & 74.50 & 67.52 & \textbf{78.42} \\
C4 & 44.34 & 46.54 & \textbf{52.28} & 46.51 & 44.82 & 50.09 & 44.54 & \uline{51.14} \\
C5 & 82.05 & 85.73 & 87.55 & 86.30 & 87.54 & \uline{88.75} & 84.98 & \textbf{88.76} \\
C6 & 69.15 & 74.21 & \uline{79.36} & 73.31 & 75.28 & 76.42 & 62.62 & \textbf{81.59} \\
C7 & \textbf{88.48} & 80.33 & 84.77 & 85.42 & \uline{86.74} & 85.64 & 84.79 & 86.64 \\
C8 & \textbf{81.65} & 67.66 & 79.66 & 68.53 & 73.20 & \uline{81.62} & 77.12 & 80.17 \\
OA & 82.01 & 84.47 & 85.58 & 84.07 & 85.79 & \uline{87.60} & 85.46 & \textbf{89.72} \\
AA & 76.05 & 75.18 & 79.84 & 75.96 & 77.60 & \uline{80.14} & 75.41 & \textbf{81.84} \\
Kappa & 74.93 & 77.94 & 79.69 & 77.43 & 79.82 & \uline{82.35} & 79.26 & \textbf{85.17} \\
mIoU & 53.85 & 57.75 & 60.52 & 57.39 & 59.27 & \uline{62.92} & 58.02 & \textbf{67.44} \\

\bottomrule
  \multicolumn{9}{p{1.1\textwidth}}{* Class labels correspond to: C1 Paddy field, C2 Dry farmland, C3 Forest land, C4 Grassland, C5 Building, C6 Highway, C7 Greenhouse, C8 Water body.} \\
  \end{tabular}
}
\end{table}

\subsection{Hyperspectral change detection}

Hyperspectral change detection (HCD) is the task of learning from a pair of image patches from two different times to predict whether the LULC of the central pixel has changed. Due to the limitations in dataset scale for HCD, the task is conventionally framed as a patch-based, pixel-wise classification problem.

\subsubsection{Datasets and experimental settings}

The HCD experiments utilized two renowned airborne change detection datasets: Bay Area and Santa Barbara. Both datasets were acquired by the AVIRIS-C sensor, providing 224 spectral bands. The Bay Area dataset was captured in 2013 and 2015 with a size of $600\times500$ pixels, while the Santa Barbara dataset was captured in 2013 and 2014 with a size of $984\times740$ pixels.

\begin{figure}[tbp]
\centering
\includegraphics[width=0.75\linewidth]{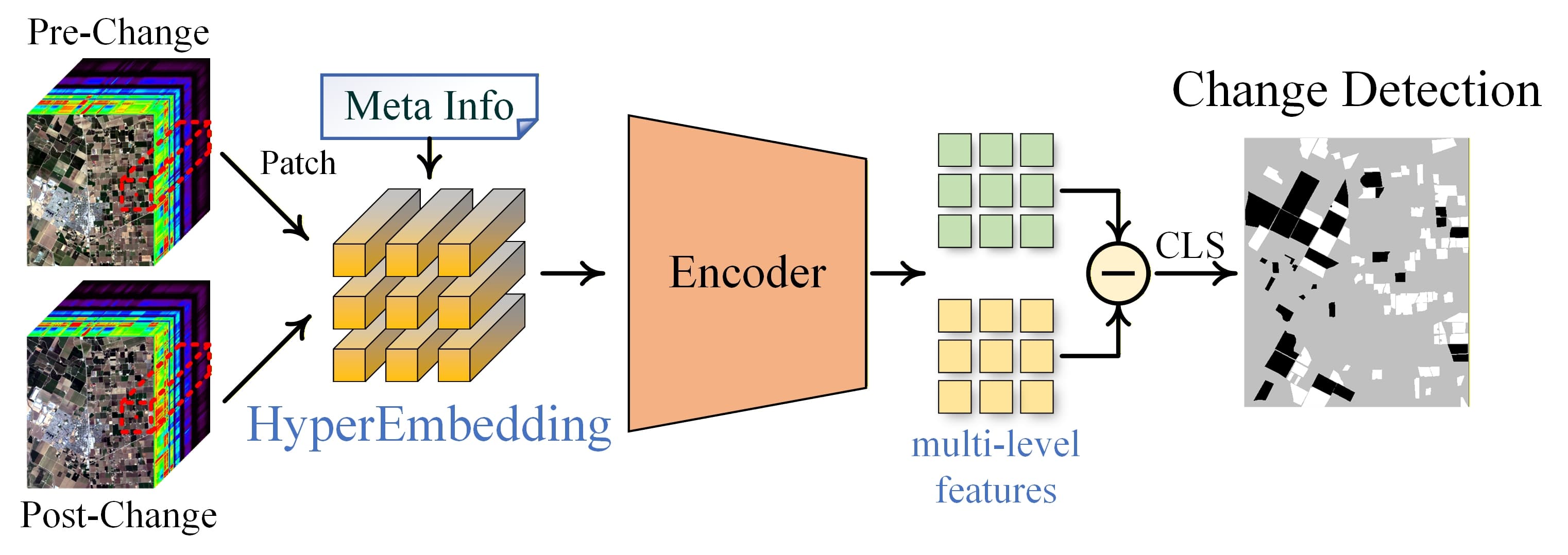}
\caption{The fine-tuning framework of SpecAware for the change detection task.}
\label{fig:CD}
\end{figure}

For the HCD task, through a simple Siamese architecture, as shown in Figure~\ref{fig:CD}, we integrated the pre-trained SpecAware model as the feature extraction backbone into the change detection framework proposed by HyperSIGMA \citep{wangHypersigmaHyperspectralIntelligence2025}. Following the settings of SST-Former \citep{wangSpectralSpatialTemporal2022}, we randomly selected 500 changed and 500 unchanged samples for  training. In the experiments, the input image size was set to 15, and the patch size of the HSI token was set as 3. The proposed method was trained for 50 epochs with a batch size of 32. The initial learning rate was set to 0.00006 and adjusted according to a cosine annealing schedule.

A comparative analysis was conducted with a wide range of established techniques, including CSANet \citep{songCSANetCrossTemporalInteraction2022}, 
SST-Former \citep{wangSpectralSpatialTemporal2022}, 
ML-EDAN \citep{quMultilevelEncoderDecoder2022}, 
GLAFormer \citep{wangGlobalLocalAttentionbased2024}
and GlobalMind \citep{huGlobalMindGlobalMultihead2024}, 
along with the other HSI pre-trained models like 
HyperSL \citep{kongHyperSLSpectralFoundation2025},
SpectralEarth \citep{brahamSpectralEarthTrainingHyperspectral2025} and 
HyperSIGMA \citep{wangHypersigmaHyperspectralIntelligence2025}. The model performance for the HCD tasks is evaluated here using the OA, Kappa, F1-score, precision, and recall.

\subsubsection{Results and analysis}

Table \ref{tab:CD_Bay} and \ref{tab:CD_Barbara} present the comprehensive evaluation results on the Bay Area and Santa Barbara datasets, respectively. As the HCD task focuses on binary classification at a fine-grained scale, our proposed SpecAware model yields competitive results on both datasets. It achieves the highest OA of 99.05\% and 99.52\% for the Bay Area and Santa Barbara datasets, respectively, and also demonstrates a performance balance between precision and recall, resulting in top F1-scores of 99.11\% and 99.39\%. Similarly, HyperSIGMA and SpectralEarth also perform well, each ranking second on one dataset separately, benefiting from their HSI data pre-training. The consistent superiority of these pretrained models confirms the necessity of jointly learning spectral and spatial features, especially for HSI data characterized by a high spatial and spectral resolution.

\begin{table}[tbp]
\caption{Quantitative results of the different change detection methods on the Bay Area dataset.}
\label{tab:CD_Bay}
\footnotesize 
\centering 
\begin{tabular}{lccccc} 
\toprule
Method & OA & Kappa & F1-score & Precision & Recall \\ 
\hline
CSANet & 98.45 & 96.87 & 98.54 & 98.41 & 98.68 \\
SST-Former & 97.58 & 95.14 & 97.73 & 98.14 & 97.31 \\
ML-EDAN & 96.89 & 93.75 & 97.06 & 97.93 & 96.21 \\
GLAFormer & 98.21 &	96.41 &	98.33 &	98.11 &	98.56  \\
GlobalMind & 97.94 & 95.87 & 98.06 & \uline{98.88} & 97.25 \\
HyperSL & 97.59 & 95.15 & 97.72 & 98.50 & 96.96 \\
SpectralEarth &	98.39 &	96.77 	&	98.48 	&	\uline{98.88} &	98.10 \\
HyperSIGMA & \uline{98.88} & \uline{97.74} & \uline{98.95} & 98.69 & \textbf{99.22} \\
SpecAware & \textbf{99.05} & \textbf{98.09} & \textbf{99.11} & \textbf{99.02} & \uline{99.20} \\
\bottomrule
\end{tabular}
\end{table}

\begin{table}[tbp]
\caption{Quantitative results of the different change detection methods on the Santa Barbara dataset.}
\label{tab:CD_Barbara}
\footnotesize 
\centering
\begin{tabular}{lccccc} 
\toprule
Method & OA & Kappa & F1-score & Precision & Recall \\ 
\hline
CSANet & 99.13 & 98.16 & 98.88 & 98.99 & 98.78 \\
SST-Former & 97.76 & 95.31 & 97.15 & 97.09 & 97.22 \\
ML-EDAN & 96.81 & 93.30 & 95.92 & 96.34 & 95.52 \\
GLAFormer &	98.59 &	97.06 &	98.22 &	97.73 &	98.71  \\
GlobalMind & 98.65 & 97.17 & 98.28 & 98.54 & 98.02 \\
HyperSL & 98.57 & 97.02 & 98.19 & 97.85 & 98.54 \\
SpectralEarth &	\uline{99.37} &	\uline{98.68} &	\uline{99.20} &	99.16 &	\uline{99.24}  \\
HyperSIGMA & 99.30 & 98.54 & 99.11 & \uline{99.25} & 98.98 \\
SpecAware & \textbf{99.52} & \textbf{98.99} & \textbf{99.39} & \textbf{99.35} & \textbf{99.42} \\
\bottomrule
\end{tabular}
\end{table}

\begin{figure}[!t]
\centering
\includegraphics[width=0.92\linewidth]{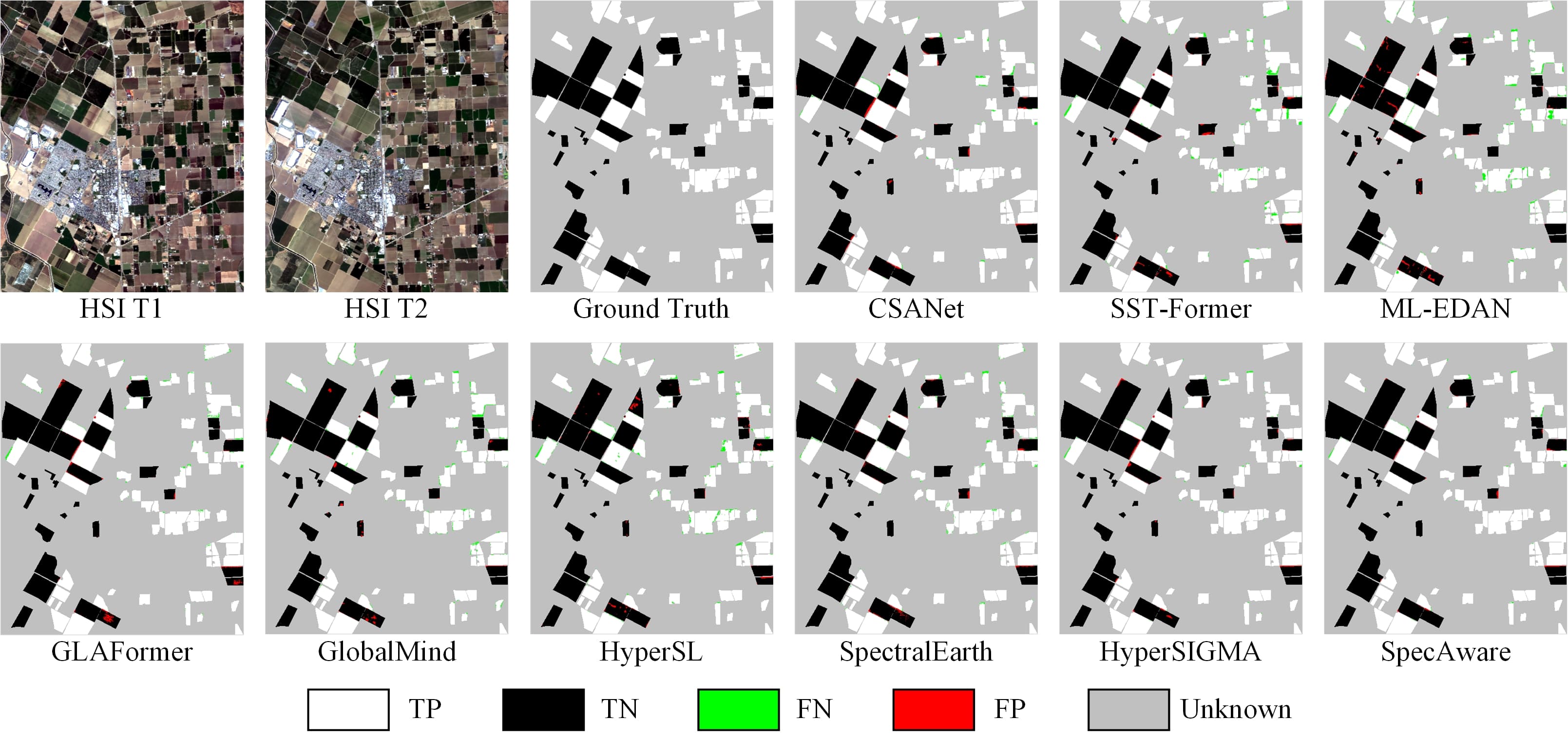}
\caption{Visual comparison of the HSI change detection results on the Bay Area dataset.}
\label{fig:CD_res1}
\end{figure}

\begin{figure}[!htb]
\centering
\includegraphics[width=0.92\linewidth]{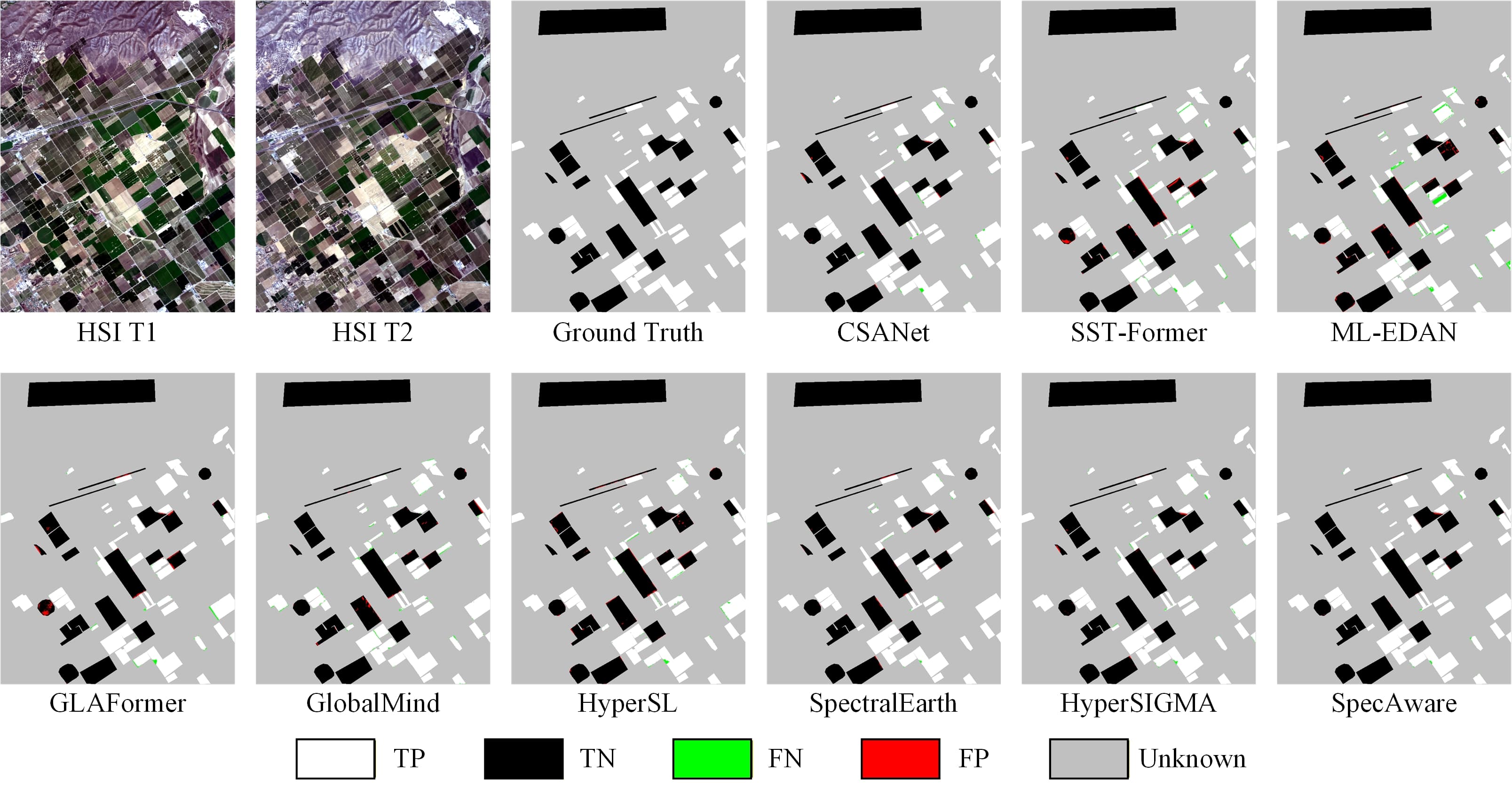}
\caption{Visual comparison of the HSI change detection results on the Santa Barbara dataset.}
\label{fig:CD_res2}
\vspace{-1.0em}
\end{figure}

Figure \ref{fig:CD_res1} and Figure~\ref{fig:CD_res2} further provide a qualitative comparison of the HCD results. Visually, the proposed SpecAware model demonstrates a superior performance. Specifically, on both datasets, SpecAware achieves the most complete detection of the true positive areas (in white), while simultaneously producing the minimal missed detections (false negatives, in green) and fewer spurious alarms (false positives, in red). This indicates a superior balance between detection integrity and noise suppression.

\subsection{Hyperspectral scene classification}

Scene classification in remote sensing is to label an entire image with a single and macroscopic LULC label (e.g., urban area, forest), unlike pixel-wise classification. The field of hyperspectral scene classification is still emerging, which plays an important role in scene positioning and scene understanding.

\subsubsection{Dataset and experimental settings}

The experiments were conducted on the HRSSC dataset \citep{liSSFNetSpectralSpatial2025}, which was derived from AVIRIS-C products. It contains 1445 images, each with a spatial size of $256\times256$ pixels, 224 spectral channels, and a ground resolution ranging from 3 to 15 m. The dataset comprises 11 classes (e.g., airport, city, farmland) and exhibits a notable class imbalance, posing a challenge to the classification task. Considering the influence of water vapor, we removed the corresponding absorption bands as well as several low-quality bands (i.e., bands 104--115, 153--170, and 223--224), ultimately retaining 192 spectral channels for the input. Following \cite{liSSFNetSpectralSpatial2025}, we used 10\% of the samples for training, and reported the best accuracy on the remaining split, evaluated every 5 epochs.

For a comprehensive comparison, the baselines included not only pre-trained hyperspectral models, i.e., HyperSIGMA \citep{wangHypersigmaHyperspectralIntelligence2025}, and SpectralEarth \citep{brahamSpectralEarthTrainingHyperspectral2025}, but also foundation models pre-trained on multispectral data modalities, i.e., SpectralGPT \citep{hongSpectralGPTSpectralRemote2024}, SatMAE \citep{congSatmaePretrainingTransformers2022}, and SatMAE++ \citep{nomanRethinkingTransformersPretraining2024}, as well as general models pre-trained on RGB data, i.e., ResNet \citep{heDeepResidualLearning2016} and ViT \citep{dosovitskiyImageWorth16x162021}.

\begin{figure}[tbp]
\centering
\includegraphics[width=0.75\linewidth]{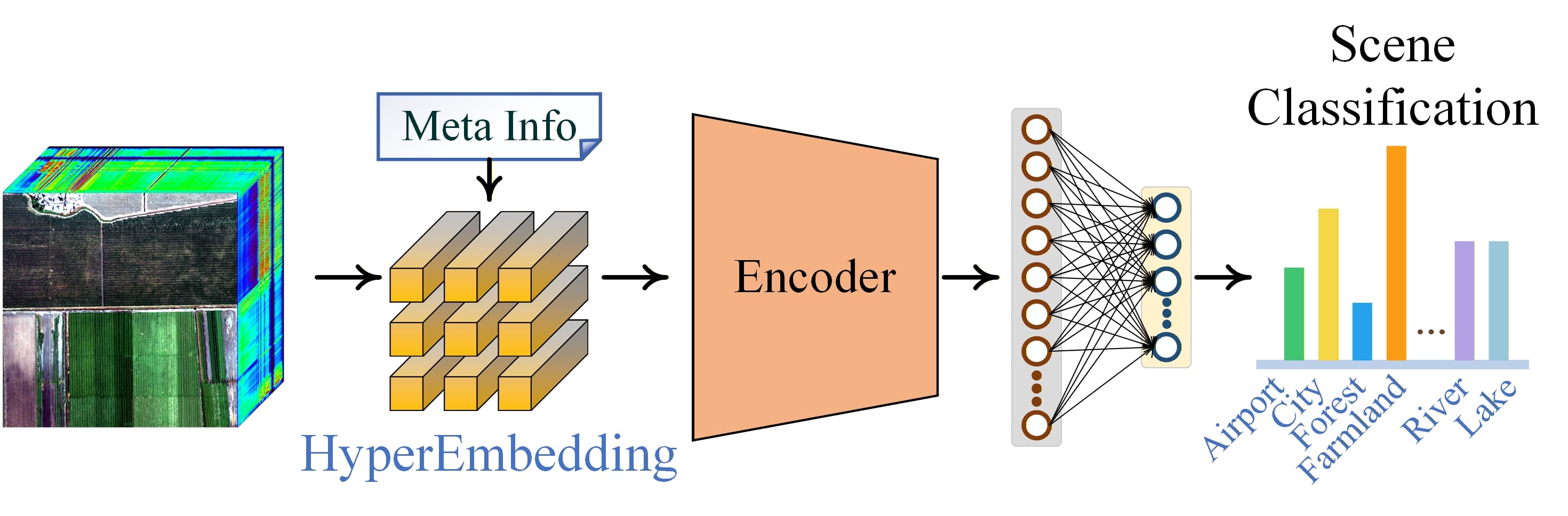}
\caption{The fine-tuning framework of SpecAware for the scene classification task.}
\label{fig:Fig_Scene}
\end{figure}

To ensure a fair comparison and adhere to the standard evaluation protocol, we adopted a uniform and minimal setup. As shown in Figure~\ref{fig:Fig_Scene}, this approach utilizes the pretrained backbone of each model to extract features and subsequently appends a single linear layer as the classifier. This simple design ensures that the performance in the scene classification task serves as an effective measure of the feature quality from the backbone, making it a robust metric for assessing pre-training effectiveness.

Due to the mismatch between the AVIRIS spectral bands and the original pre-training configuration, all the comparative models were trained using an end-to-end fine-tuning strategy, with all the parameters being trainable. Notably, both the proposed SpecAware model (via its HyperEmbedding module) and the SpectralEarth model (owing to its spectral adaptation head) are capable of processing inputs with an arbitrary number of HSI bands. This inherent flexibility prompted us to subject both models to an additional linear probing test, where their pre-trained backbones were frozen and only the linear heads were trained.

The training ran for 50 epochs using the AdamW optimizer with an initial learning rate of 0.0001 and 0.001 for the remote sensing and general-purpose foundation models, respectively. In addition, a StepLR scheduler was employed to decay the learning rate by a factor of 0.1 every 20 epochs. For data augmentation, we randomly cropped $224\times224$ patches from the $256\times256$ images during training, whereas a $224\times224$ center crop was used at test time. The performance for all the experiments is evaluated here using three standard metrics: OA, Kappa, and F1-score.

\subsubsection{Results and analysis}

\begin{figure}[!htb]
\centering
\includegraphics[width=0.65\linewidth]{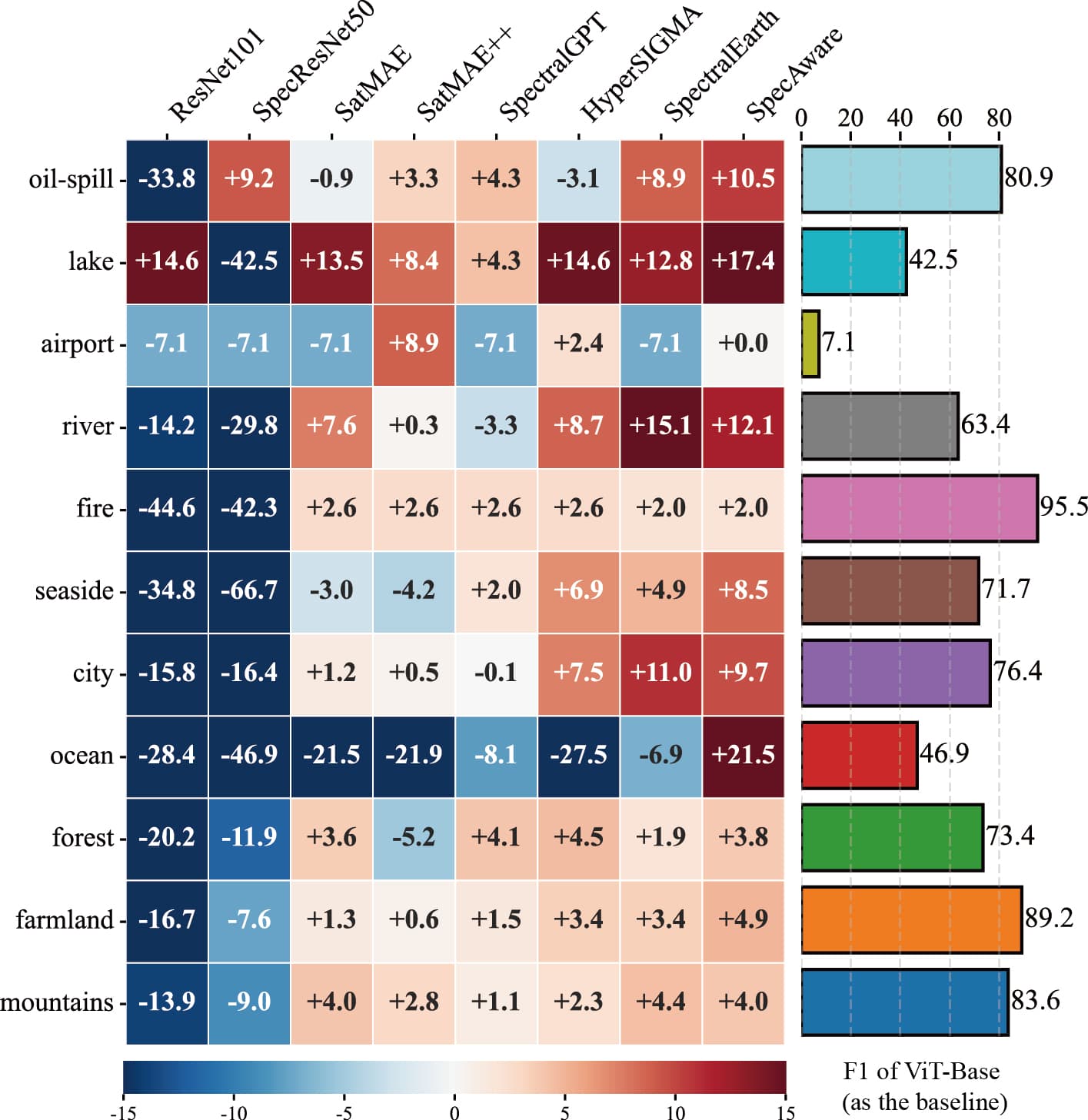}
\caption{Relative F1-score performance improvement for each class on the baseline of ViT-Base.}
\label{fig:F1}
\end{figure}

Table \ref{tab:Scene_CLS} summarizes the quantitative results of the different pre-trained models on the HRSSC scene classification dataset. The proposed SpecAware model achieves the superior performance, reaching 85.22\% in OA and 75.01\% in F1-score. This is followed by SpectralEarth and HyperSIGMA, which obtain accuracies of 84.61\% and 81.85\% in OA, and 71.01\% and 68.45\% in F1-score, respectively. Notably, the proposed SpecAware model exhibits a more pronounced advantage in terms of the F1-score, which indicates its strong generalization capabilities. Figure~\ref{fig:F1} provides a detailed per-class F1-score for each model, using the ViT-Base model as a performance baseline. The results indicate that the proposed SpecAware model achieves marked improvements across most of the classes, particularly for oil-spill, lake, ocean, and seaside. Furthermore, a notable performance hierarchy emerges from the results. The three models pre-trained on HSI data outperform those trained on multispectral remote sensing data. These multispectral pre-trained models, in turn, are better than the general-purpose models pre-trained on ImageNet (RGB), which highlights a domain gap between the multispectral/RGB and HSI feature spaces.

\begin{table}[htbp]
\caption{Quantitative results of the different pre-trained models on the HRSSC dataset for scene classification.}
\label{tab:Scene_CLS}
\footnotesize
\centering
\begin{tabular}{ccccc}
\toprule
Method & Pretraining
dataset & OA & Kappa & F1-score \\
\hline
ResNet101 & ImageNet-1K & 57.20 & 50.52 & 46.89 \\
SpecResNet50 & SpectralEarth & 66.16 & 59.62 & 41.80 \\
ViT-Base & ImageNet-1K & 78.71 & 75.16 & 66.43 \\
SatMAE & fMoW-S2 & 80.32 & 77.13 & 66.55 \\
SatMAE++ & fMoW-S2 & 79.10 & 75.67 & 66.07 \\
SpectralGPT+ & fMoW-S2+BigEarthNet & 79.86 & 76.58 & 66.55 \\
HyperSIGMA & HyperGlobal-450K & 81.85 & 78.88 & 68.45 \\
SpectralEarth* & SpectralEarth & 80.78 & 77.52 & 69.57 \\
SpectralEarth & SpectralEarth & \uline{84.61}\uline{} & \uline{82.07}\uline{} & 71.01 \\
SpecAware* & Hyper-400K & 84.00 & 81.34 & \uline{71.55}\uline{} \\
SpecAware & Hyper-400K & \textbf{85.22} & \textbf{82.86} & \textbf{75.01} \\
\bottomrule
  \multicolumn{5}{p{0.65\textwidth}}{* Denotes the linear probing protocol applied to SpectralEarth and SpecAware, where only the final linear head, consisting of 8459 trainable parameters, was trained.} \\
\end{tabular}
\end{table}

Under the linear probing protocol, where only 8459 parameters (0.009\% of the total) were optimized, the proposed SpecAware* model achieves an OA of 84.00\%. This performance suggests the effectiveness of the hypernetwork-driven HyperEmbedding framework, which encodes hyperspectral images through a channel-wise decomposition approach to enable compatibility with sensors of varying channel configurations.

\begin{figure}[!htb]
\centering
\includegraphics[width=1.0\linewidth]{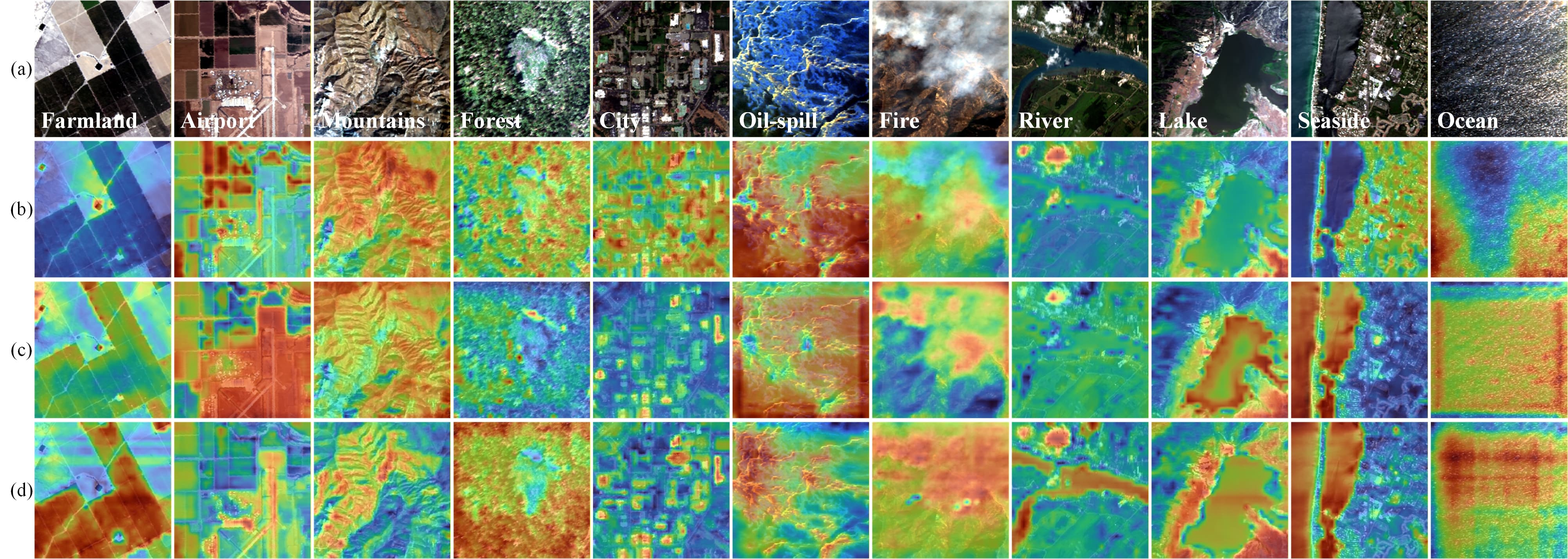}
\caption{Visualization of the feature extracted by three hyperspectral foundation models after finetuning. (a)image, (b) SpectralEarth, (c) HyperSIGMA, (d) SpecAware.}
\label{fig:heapmap}
\vspace{-0.3em}
\end{figure}

Figure~\ref{fig:heapmap} presents the feature maps extracted from the last backbone layer of the three hyperspectral foundation models after finetuning. These visualizations show feature maps rather than gradient-based attribution; therefore, we focus on the boundaries and object structures across land cover categories. Compared with the HyperSIGMA and SpectralEarth models, the proposed SpecAware model more effectively distinguishes different land-cover semantics and captures high-quality boundaries (e.g., farmland areas, lake edges and building layouts). This demonstrates its sophisticated understanding of LULC, leveraging both fine-grained local structures and holistic scene context to encode discriminative class-specific textures.

\section{Discussion}

\subsection{Further Experiments}

\subsubsection{Model Scalability.}

We further investigate the effect of scaling SpecAware to larger model capacities. As reported in Table~\ref{tab:Scale_experiments}, SpecAware-B (ViT-Base, 86M) and SpecAware-L (ViT-Large, 307M) were evaluated across semantic segmentation and scene classification tasks, along with two others hyperspectral foundation models. Specifically, SpecAware-L achieves an mIoU of 68.52 on the WHU-H\textsuperscript{2}SR dataset, representing an improvement of 1.08 over the Base version; and its scene classification F1-score also increases from 75.01 to 76.12. These results indicate that scaling up the model allows SpecAware to more deeply exploit the rich spatial-spectral knowledge embedded in the large Hyper-400K dataset.

\begin{table}[htbp]
\caption{Performance comparison of ViT-Base and Large variants of the three hyperspectral foundation models.}
\label{tab:Scale_experiments}
\footnotesize
\centering

\begin{tabular}{ccccc}
\toprule
\multirow{2}{*}{Methods} 
    & \multicolumn{2}{c}{WHU-H\textsuperscript{2}SR} 
    & \multicolumn{2}{c}{HRSSC} \\
\cmidrule(lr){2-3} \cmidrule(lr){4-5}
    & OA & mIoU & OA & F1-score \\
\midrule
HyperSIGMA-B        & 85.46 & 58.02 & 81.85 & 68.45 \\
SpectralEarth-B     & 87.60 & 62.92 & 84.61 & 71.01 \\
SpecAware-B         & \textbf{89.72} & \textbf{67.44} & \textbf{85.22} & \textbf{75.01} \\
HyperSIGMA-L        & 85.75 & 58.25 & 81.85 & 69.40 \\
SpectralEarth-L     & 88.64 & 64.81 & 82.62 & 70.76 \\
SpecAware-L         & \textbf{90.10} & \textbf{68.52} & \textbf{86.22} & \textbf{76.12} \\
\bottomrule
\end{tabular}
\end{table}

In addition, both SpecAware-B and SpecAware-L outperform HyperSIGMA and SpectralEarth under comparable model size. We attribute this to the proposed HyperEmbedding module, that enhances adaptability to diverse spectral channel configurations across heterogeneous hyperspectral sensors. These results indicate that larger-capacity pretrained models remain essential for capturing complex spatial-spectral structures in high-dimensional HSI data. It is worth noting that scaling up the model may lead to overfitting and result in degraded performance, a phenomenon also reported in previous studies, e.g., SpectralEarth \citep{brahamSpectralEarthTrainingHyperspectral2025} and SAR-JEPA \citep{liPredictingGradientBetter2024}. Therefore, stronger regularization strategies, e.g., stochastic depth \citep{huangDeepNetworksStochastic2016a}, are required to maintain stable performance.

\subsubsection{Transfer to satellite sensors.}

\begin{figure}[!htb]
\centering
\includegraphics[width=1.0\linewidth]{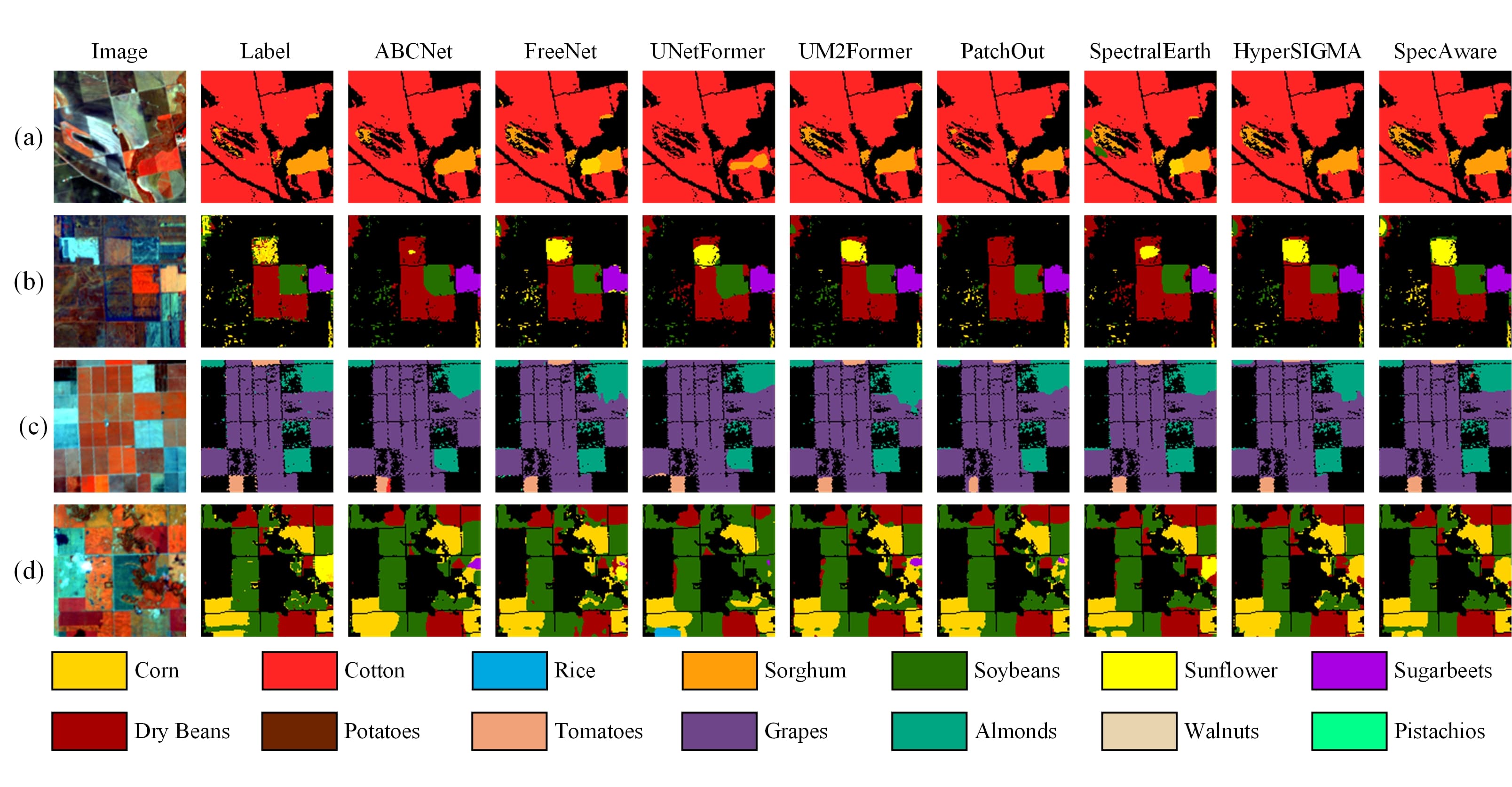}
\caption{Visual comparison of the HSI data land-cover segmentation results of the different methods on the EO1-CDL dataset.}
\label{fig:seg_eo1}
\vspace{-0.3em}
\end{figure}

To evaluate the generalization of the proposed SpecAware model to satellite-based hyperspectral sensors, we conducted experiments on the EO1-CDL dataset \citep{brahamSpectralEarthTrainingHyperspectral2025}. This dataset was acquired by the EO-1 Hyperion sensor, which contains 198 valid spectral bands in calibrated radiance format, with a spatial resolution of 30\,m. The EO1-CDL dataset comprises 14 agricultural categories. As presented in Table 10, SpecAware achieves the superior performance among all compared methods, with an overall accuracy of 80.37\% and an mIoU of 52.13\%. The corresponding visualization results are shown in Figure~\ref{fig:seg_eo1}. These results show that despite difference in spatial resolution and imaging characteristics, the pretrained SpecAware model demonstrates extensibility to satellite hyperspectral sensors.

\begin{table}[htbp]
\caption{Quantitative results of the different methods on the EO1-CDL dataset.}
\label{tab:Seg_EO1}
\footnotesize
\centering

\begin{tabular}{ccccc}
\toprule
Method        & OA     & AA     & Kappa  & mIoU  \\
\hline
ABCNet        & 75.19  & 57.30  & 70.18  & 46.16 \\
FreeNet       & 77.82  & 60.65  & 73.37  & 47.62 \\
UNetFormer    & 75.44  & 56.08  & 70.54  & 44.14 \\
UM2Former     & 77.65  & 60.58  & 73.22  & 47.50 \\
PatchOut      & 78.23  & 59.47  & 73.85  & 48.08 \\
SpectralEarth & \uline{80.04} & \uline{63.22} & \uline{76.09} & \uline{51.19} \\
HyperSIGMA    & 78.72  & 60.15  & 74.45  & 48.33 \\
SpecAware     & \textbf{80.37} & \textbf{64.13} & \textbf{76.45} & \textbf{52.13} \\
\bottomrule
\end{tabular}
\end{table}

\subsection{Ablation Study}

In this section, we present a series of ablation studies to evaluate the contribution of each major component in the proposed framework, including the hypernetwork input conditions, the meta info condition tokens, the embedding methods for HSI tokens, the stage-based pretraining approach and loss functions. All ablation experiments, except for group (d), were pretrained for 200 epochs on the AVIRIS-3 L1 subset and further finetuned for 50 epochs on the multi-sensor Hyper-90K subset, followed by linear-probe evaluation on the downstream scene classification task. The experimental results are summarized in Table~\ref{tab:Ablation_res}. The last row (SpecAware) corresponds to the complete configuration and serves as the full-model reference. For each ablation group, all experiments are performed by removing individual components from the full model to isolate their effects.

\vspace{0.0em}
\begin{table}[htb]
\caption{Ablation study of SpecAware Components using linear-probe evaluation on the HRSSC dataset.}
\label{tab:Ablation_res}
\centering
\footnotesize

\begin{tabular}{lcc}
\toprule
Component / Condition & OA & F1-score \\
\midrule

(a) \textbf{Hypernetwork input conditions} & & \\
\quad\quad Meta info conditions only  & 81.85 & 69.58 \\
\quad\quad Content conditions only     & 65.77 & 52.73 \\
\midrule

(b) \textbf{Meta info conditions tokens} & & \\
\quad\quad Wavelength encoding only   & 80.09 & 67.37 \\
\quad\quad Wavelength + FWHM          & 82.01 & 68.74 \\
\midrule

(c) \textbf{Embedding method for HSI token} & & \\
\quad\quad Convolution weight generation & 79.56 & 66.86 \\
\midrule

(d) \textbf{Stage-based pretraining} & & \\
\quad\quad Hyper-400K directly & 81.55 & 68.66 \\
\midrule

(e) \textbf{Loss functions} & & \\
\quad\quad MSE Loss          & 81.85 & 69.32 \\
\quad\quad Charbonnier Loss  & 82.39 & 70.35 \\
\midrule

\textbf{SpecAware (full configuration)} & 83.15 & 70.71 \\
\bottomrule

\end{tabular}
\end{table}

(a) \textbf{Meta info and content-aware guidance module.} As shown in Table~\ref{tab:Ablation_res}, the attribute-only variant and the content-only variant achieve the accuracy of 81.85\% and 65.77\% in terms of OA, respectively, which indicates the importance of meta-attribute features to the hypernetwork. Then, incorporating both the meta info-based features and content-aware features further improves the OA from 81.85\% to 83.15\%. Different from the standard MAE approaches that rely primarily on spatial redundancy, HSI pre-training must also capture the distinctive spatial-spectral signatures. By conditioning the hypernetwork on meta-content features, our framework generates patch-specific feature extractors that adapt to the spectral characteristics of different scenes. In this study, simple spectral statistics are employed as the content-aware guidance. As LLMs have shown benefits for guiding downstream tasks, future work will investigate their integration with HSI data.

(b) \textbf{Meta info aware guidance token}. We further evaluated the effect of the meta-info conditions. Using wavelength alone yields an OA of 80.09\%, while incorporating FWHM increases the accuracy to 82.01\%, and adding dataset-level attributes (sensor name and data-processing level) further improves OA to 83.15\%. This trend is consistent with findings from Copernicus-FM \citep{wangUnifiedCopernicusFoundation2025}, where introducing FWHM and sensor attributes enhances model generalization. Under multi-sensor pre-training, richer attribute information enables the network to more effectively capture both similarities and differences among sensors. Our utilization of LLM-based attribute encoding aims to better support future meta info scalability and cross-sensor/modal compatibility.

(c) \textbf{Channel-wise decomposition.} We compared the proposed channel-wise spectral decomposition with the convolution weight generation approach (DOFA-style). Although both methods employ hypernetwork-generated weights and, in principle, can handle arbitrary spectral dimensionality, the conventional formulation shares kernel parameters across samples within a batch, thereby limiting the incorporation of sample-specific content guidance. Under comparable settings, the channel-wise formulation (without content guidance) achieves an OA of 81.85\%, while the DOFA-style convolution generation achieves 79.56\%. This result indicates that channel-wise decomposition is more suitable for HSI data with large variability in spectral channel configurations.

(d) \textbf{Stage-based pre-training.} Compared with training directly on the full Hyper-400K dataset, with an equivalent number of pretraining optimization steps, the staged procedure yields an improvement of 1.6 percentage points in OA. This result suggests that progression from simpler to more complex spectral conditions is advantageous for HSI representation learning. Further investigation into more efficient SSL paradigms will be pursued in future work.

(e) \textbf{Multiple loss functions.} We first compare against the MSE loss commonly used in MAE-based frameworks. Using MSE alone yields an OA of 81.85\%, while replacing it with the Charbonnier loss increases OA to 82.39\%, reflecting stronger robustness to noise in HSI data. Then, introducing the SAM loss further improves OA by 0.76 percentage points. This result reflects the importance of enforcing spectral angular consistency, which helps preserve the spectral shape and is essential for reliable material discrimination in hyperspectral SSL.

\subsection{Model parameter analysis}

\textbf{Embedding module parameters.} We analyze the parameters sizes of the image embedding layers in the three hyperspectral foundation models. HyperSIGMA uses a vanilla convolution embedding layer whose parameters grow linearly with the number of input channels (e.g., 4.9M for 100 channels). SpectralEarth introduces a spectral adapter that reduces the spectral dimension to 128 before a standard convolution embedding, resulting in about 6.3M parameters. The proposed hypernetwork-driven HyperEmbedding module contains 4.4M parameters (3.6M in the hypernetwork) and does not expand with increasing spectral channels, making it a compact and efficient embedding design relative to the full ViT architecture.

\textbf{Full model parameters.} Table~\ref{tab:flops} summarizes the parameter sizes of different semantic segmentation models. Generally, the task-specific models have fewer parameters than foundation models. However, for complex hyperspectral tasks, e.g., semantic segmentation, larger model capacity is often required to learn more expressive and transferable spatial-spectral representations, enabling better generalization across diverse scenes and sensors. For foundation models, as both SpecAware and SpectralEarth are built upon a standard ViT architecture, they exhibit similar parameter size, while HyperSIGMA introduces separate spectral and spatial ViT branches, resulting in nearly twice the parameters. Besides, it is also worth noting that the scale of hyperspectral pre-training models remains considerably smaller than in computer vision or multispectral remote sensing, underscoring the need for more efficient and scalable hyperspectral foundation architectures, e.g., mixture-of-experts (MOE) designs \citep{riquelmeScalingVisionSparse2021}.

\begin{table}[htb]
\caption{Total trainable parameters and FLOPs for the different methods on the semantic segmentation tasks.}
\label{tab:flops}
\centering
\scriptsize
\makebox[\textwidth][c]{
\begin{tabular}{C{2.1cm} *{8}{C{1.6cm}}}
\toprule
Method & ABCNet & FreeNet & UNetFormer & UM2Former & PatchOut & SpectralEarth & HyperSIGMA & SpecAware \\
\midrule
Params & 14.97\,M & 2.84\,M & 12.45\,M & 6.99\,M & 30.79\,M & 96.33\,M & 237.04\,M & 93.55\,M \\
FLOPs  & 29.65\,G & 41.29\,G & 15.63\,G & 13.67\,G & 69.11\,G & 165.26\,G & 146.54\,G & 155.91\,G \\
\bottomrule
\end{tabular}
}
\end{table}

\section{Conclusion}

In this paper, we have proposed SpecAware, an adaptive pre-training framework for multi-sensor airborne HSI data, driven by a meta-feature conditioned hypernetwork, to enable generalization capabilities across varied sensors, spectral ranges, and data levels in HSI LULC mapping tasks. We also constructed a large-scale, multi-sensor, multi-level aerial HSI pre-training dataset to support this research. The core innovation of SpecAware lies in its meta-content aware encoder and adaptive hypernetwork HyperEmbedding architecture, which enables unified joint pre-training on HSI data from diverse sensors with variable spectral bands. Driven jointly by spectral and sensor attributes and the intrinsic content features in the imagery, the hypernetwork learns to extract the spatial-spectral characteristics for each spectral channel of different scenes, and also to bridge the gap between radiance and reflectance information. To accelerate the pre-training on massive data, we also introduced a multi-view distributed pre-training strategy, which allows the model to learn highly generalizable features while maintaining computational efficiency. Experiments on LULC downstream tasks, including large-scale land-cover semantic segmentation, change detection, and scene classification, demonstrated that the representations learned by SpecAware outperformed the existing pre-trained and popular fully supervised models, while exhibiting a strong transfer performance. We believe that the proposed SpecAware framework will provide a novel paradigm and a foundation resource for the advancement of HSI foundation models.

However, despite the large-scale multi-sensor Hyper-400K dataset we have collected, fully realizing the potential of SSL for HSI still requires expanding to additional sensors, particularly through the integration of satellite and airborne imagery. Moreover, SSL for HSI also remains computationally demanding, as pixel-level spectral reconstruction is far more complex than RGB reconstruction, and is further impacted by spectral redundancy and noise. In future work, we plan to refine SSL frameworks by incorporating MoE \citep{riquelmeScalingVisionSparse2021} and physics-informed mechanisms \citep{tangPhySwinEfficientPhysicallyInformed2025} to enhance model capacity and provide high-quality pretraining weights for a wider range of downstream HSI tasks.

\section*{Data availability}
All data used to construct the pre-training dataset were obtained from the official AVIRIS data portal. To support reproducibility, we will release the corresponding data inventory and preprocessing pipeline.

\section*{CRediT authorship contribution statement}

\textbf{Renjie Ji:} Writing -- Original Draft, Writing -- Review \& Editing, Methodology, Software, Validation, Investigation, Data Curation, Visualization. \textbf{Xue Wang:} Methodology, Software, Validation, Resources, Writing -- Original Draft, Writing -- Review \& Editing. \textbf{Chao Niu:} Methodology, Investigation, Data Curation, Validation. \textbf{Wen Zhang:} Data Curation, Investigation, Resources. \textbf{Yong Mei:} Investigation, Resources. \textbf{Kun Tan:} Conceptualization, Writing -- Original Draft, Writing -- Review \& Editing, Supervision, Funding acquisition.

\section*{Declaration of competing interest}

The authors declare that they have no known competing financial interests or personal relationships that could have appeared to influence the work reported in this paper.

\section*{Acknowledgement}

We would like to express our sincere gratitude to the NASA Jet Propulsion Laboratory for their sustained efforts in collecting and sharing a vast amount of airborne HSI data. This work was supported in part by the Yangtze River Delta Science and Technology Innovation Community Joint Research (Basic Research) Project (No. 2024CSJZN1300), the Shanghai Municipal Education Commission Science and Technology Project (2024AI02002), the National Natural Science Foundation of China (No. 42171335), the National Civil Aerospace Project of China (No. D040102), and the China Postdoctoral Science Foundation (2024M760925, 2025T180089).

\bibliographystyle{elsarticle-harv}
\bibliography{v2}

\end{document}